\documentclass[UTF8]{article} 
\usepackage{iclr2023_conference,times}

\usepackage{amsmath,amsfonts,bm}

\def\eqref#1{equation~\ref{#1}}

\def\1{\bm{1}}

\DeclareMathAlphabet{\mathsfit}{\encodingdefault}{\sfdefault}{m}{sl}
\SetMathAlphabet{\mathsfit}{bold}{\encodingdefault}{\sfdefault}{bx}{n}

\usepackage{microtype}
\usepackage{amsmath}
\usepackage{booktabs}
\usepackage{colortbl}
\usepackage{CJKutf8}
\usepackage[utf8]{inputenc}
\definecolor{lightgray}{rgb}{0.9,0.9,0.9}
\usepackage{caption}
\usepackage{subcaption}
\usepackage{xcolor}
\usepackage{graphicx}
\usepackage{setspace}
\usepackage{hyperref}
\usepackage{url}
\usepackage{multirow}
\usepackage{colortbl}
\usepackage{tabularx}
\usepackage{blindtext}
\usepackage{pgfplots}
\pgfplotsset{compat=1.18} 
\usepackage{tikz}
\usetikzlibrary{er,positioning,bayesnet}
\usepackage[inline]{enumitem}
\usepackage{makecell}
\usepackage{tipa}
\usepackage{siunitx}
\usepackage{tocloft}
\usepackage{listings}
\usepackage[raster,skins]{tcolorbox} 
\usepackage{xltabular}
\usepackage[framemethod=tikz]{mdframed}
\surroundwithmdframed[
  hidealllines=true,
  innerleftmargin=0pt,
  innertopmargin=0pt,
  innerbottommargin=0pt]{lstlisting}
\lstnewenvironment{response}[1][] 
 {\lstset{
 columns=fullflexible,
  breakautoindent=false, breakindent=0pt, breaklines, linewidth=8cm, #1}}
 {}
\tcbuselibrary{breakable}

\title{Qwen Technical Report}
\author{
\\
\parbox{\linewidth}{Jinze Bai, Shuai Bai, Yunfei Chu, Zeyu Cui, Kai Dang, Xiaodong Deng, Yang Fan, Wenbin Ge, Yu Han, Fei Huang, Binyuan Hui, Luo Ji, Mei Li, Junyang Lin, Runji Lin, Dayiheng Liu, Gao Liu, Chengqiang Lu, Keming Lu, Jianxin Ma, Rui Men, Xingzhang Ren, Xuancheng Ren, Chuanqi Tan, Sinan Tan, Jianhong Tu, Peng Wang, Shijie Wang, Wei Wang, Shengguang Wu, Benfeng Xu, Jin Xu, An Yang, Hao Yang, Jian Yang, Shusheng Yang, Yang Yao, Bowen Yu, Hongyi Yuan, Zheng Yuan, Jianwei Zhang, Xingxuan Zhang, Yichang Zhang, Zhenru Zhang, Chang Zhou, Jingren Zhou, Xiaohuan Zhou, Tianhang Zhu.}
\AND
Qwen Team, Alibaba Group\thanks{Authors are ordered alphabetically by the last name. Correspondence to: ericzhou.zc@alibaba-inc.com.}   \\
}

\newcommand{\qwen}{\textsc{Qwen}}
\newcommand{\qwenchat}{\textsc{Qwen-Chat}}

\newcommand{\qwensevenbchat}{\textsc{Qwen-7B-Chat}}
\newcommand{\qwenfourteenbchat}{\textsc{Qwen-14B-Chat}}
\newcommand{\qwencoder}{\textsc{Code-Qwen}}
\newcommand{\qwencodersevenb}{\textsc{Code-Qwen-7B}}
\newcommand{\qwencoderfourteenb}{\textsc{Code-Qwen-14B}}
\newcommand{\qwencoderchat}{\textsc{Code-Qwen-Chat}}
\newcommand{\qwencodersevenbchat}{\textsc{Code-Qwen-7B-Chat}}
\newcommand{\qwencoderfourteenbchat}{\textsc{Code-Qwen-14B-Chat}}
\newcommand{\mathqwen}{\textsc{Math-Qwen}}
\newcommand{\mathqwenchat}{\textsc{Math-Qwen-Chat}}
\newcommand{\mathqwensevenbchat}{\textsc{Math-Qwen-7B-Chat}}
\newcommand{\mathqwenfourteenbchat}{\textsc{Math-Qwen-14B-Chat}}
\newcommand{\qwenvl}{\textsc{Qwen-VL}}
\newcommand{\qwenvlchat}{\textsc{Qwen-VL-Chat}}

\newcommand{\llamatwo}{\textsc{Llama 2}}
\newcommand{\llamatwochat}{\textsc{Llama 2-Chat}}
\newcommand{\codellama}{\textsc{Code LLaMA}}
\iclrfinalcopy
\begin{document}

\maketitle

\begin{abstract}
Large language models (LLMs) have revolutionized the field of artificial intelligence, enabling natural language processing tasks that were previously thought to be exclusive to humans. In this work, we introduce \qwen{}\footnote{\qwen{} is a moniker of Qianwen, which means ``thousands of prompts'' in Chinese. The pronunciation of ``\qwen{}'' can vary depending on the context and the individual speaking it. Here is one possible way to pronounce it: /kw\textipa{E}n/.}, the first installment of our large language model series. \qwen{} is a comprehensive language model series that encompasses distinct models with varying parameter counts. It  includes \qwen{}, the base pretrained language models, and \qwenchat{}, the chat models finetuned with human alignment techniques. 
The base language models consistently demonstrate superior performance across a multitude of downstream tasks, and the chat models, particularly those trained using Reinforcement Learning from Human Feedback (RLHF), are highly competitive. 
The chat models possess advanced tool-use and planning capabilities for creating agent applications, showcasing impressive performance even when compared to bigger models on complex tasks like utilizing a code interpreter.
Furthermore, we have developed coding-specialized models, \qwencoder{} and \qwencoderchat{}, as well as mathematics-focused models, \mathqwenchat{}, which are built upon base language models. 
These models demonstrate significantly improved performance in comparison with open-source models, and slightly fall behind the proprietary models. 
  
\end{abstract}
\clearpage

\tableofcontents
\clearpage

\section{Introduction}
\label{sec:intro}

Large language models (LLMs)~\citep{gpt, bert, t5, gpt3, gpt4, palm, palm2, LaMDA, llama, llama2} have revolutionized the field of artificial intelligence (AI) by providing a powerful foundation for complex reasoning and problem-solving tasks. These models have the ability to compress vast knowledge into neural networks, making them incredibly versatile agents. With a chat interface, LLMs can perform tasks that were previously thought to be the exclusive domain of humans, especially those involving creativity and expertise~\citep{chatgpt,instructgpt,palm2,bard,claude,claude2}. They can engage in natural language conversations with humans, answering questions, providing information, and even generating creative content such as stories, poems, and music. This has led to the development of a wide range of applications, from chatbots and virtual assistants to language translation and summarization tools.

LLMs are not just limited to language tasks. They can also function as a generalist agent~\citep{generalist,ofasys, voyager, autogpt, metagpt}, collaborating with external systems, tools, and models to achieve the objectives set by humans. For example, LLMs can understand multimodal instructions~\citep{gpt4, qwenvl, llava, mplug-owl, instructblip, kosmos2}, execute code~\citep{codex, CodeGeeX, starcoder}, use tools~\citep{toolformer, langchain, autogpt}, and more. 
This opens up a whole new world of possibilities for AI applications, from autonomous vehicles and robotics to healthcare and finance.
As these models continue to evolve and improve, we can expect to see even more innovative and exciting applications in the years to come. Whether it's helping us solve complex problems, creating new forms of entertainment, or transforming the way we live and work, LLMs are poised to play a central role in shaping the future of AI.

\begin{figure}[htbp]
\centering
\includegraphics[width=0.9\textwidth]{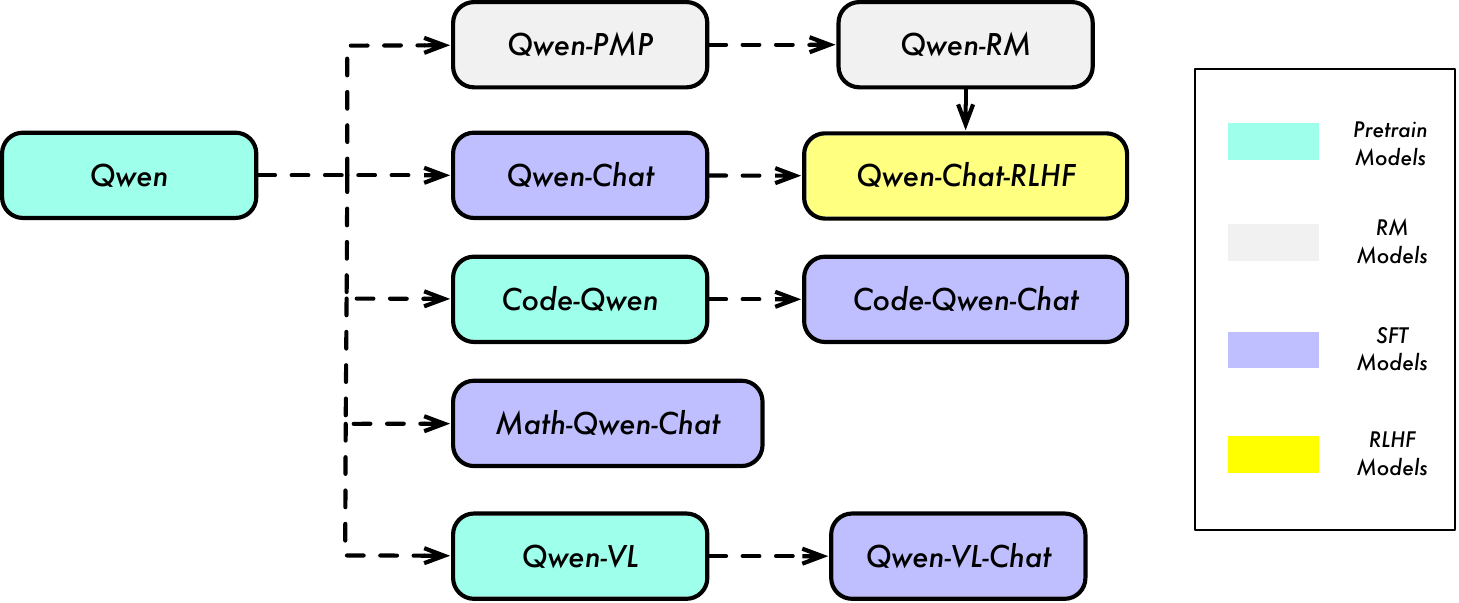}
\caption{\textbf{Model Lineage of the Qwen Series.} We have pretrained the language models, namely \qwen{}, on massive datasets containing trillions of tokens. We then use SFT and RLHF to align \qwen{} to human preference and thus we have \qwenchat{} and specifically its improved version \qwenchat{}-\textsc{RLHF}. Additionally, we also develop specialized models for coding and mathematics, such as \qwencoder{}, \qwencoderchat{}, and \mathqwenchat{} based on \qwen{} with similar techniques. Note that we previously released the multimodal LLM, \textsc{Qwen-VL} and \textsc{Qwen-VL-Chat}~\citep{qwenvl}, which are also based on our \qwen{} base models.}
\label{fig:overview}
\end{figure}

Despite their impressive capabilities, LLMs are often criticized for their lack of reproducibility, steerability, and accessibility to service providers. 
In this work, we are pleased to present and release the initial version of our LLM series, \qwen{}. 
\qwen{} is a moniker that derives from the Chinese phrase Qianwen, which translates to ``thousands of prompts'' and conveys the notion of embracing a wide range of inquiries. 
\qwen{} is a comprehensive language model series that encompasses distinct models with varying parameter counts. The model series include the base pretrained language models, chat models finetuned with human alignment techniques, i.e., supervised finetuning (SFT), reinforcement learning with human feedback (RLHF), etc., as well as specialized models in coding and math. The details are outlined below:

\begin{enumerate}
\item 
The base language models, namely \qwen{}, have undergone extensive training using up to \num{3} trillion tokens of diverse texts and codes, encompassing a wide range of areas. These models have consistently demonstrated superior performance across a multitude of downstream tasks, even when compared to their more significantly larger counterparts.

\item 
The \qwenchat{} models have been carefully finetuned on a curated dataset relevant to task performing, chat, tool use, agent, safety, etc. The benchmark evaluation demonstrates that the SFT models can achieve superior performance. Furthermore, we have trained reward models to mimic human preference and applied them in RLHF for chat models that can produce responses preferred by humans. Through the human evaluation of a challenging test, we find that \qwenchat{} models trained with RLHF are highly competitive, still falling behind GPT-4 on our benchmark. 

\item In addition, we present specialized models called \qwencoder{}, which includes \qwencodersevenb{} and \qwencoderfourteenb{}, as well as their chat models, \qwencoderfourteenbchat{} and \qwencodersevenbchat{}. Specifically, \qwencoder{} has been pre-trained on extensive datasets of code and further fine-tuned to handle conversations related to code generation, debugging, and interpretation. The results of experiments conducted on benchmark datasets, such as HumanEval~\citep{codex}, MBPP~\citep{mbpp}, and HumanEvalPack~\citep{octopack}, demonstrate the high level of proficiency of \qwencoder{} in code understanding and generation. 

\item This research additionally introduces \mathqwenchat{} specifically designed to tackle mathematical problems.
Our results show that both \mathqwensevenbchat{} and \mathqwenfourteenbchat{} outperform open-sourced models in the same sizes with large margins and are approaching GPT-3.5 on math-related benchmark datasets such as GSM8K~\citep{gsm8k} and MATH~\citep{hendrycks2021measuring}.

\item Besides, we have open-sourced \qwenvl{} and \qwenvlchat{}, which have the versatile ability to comprehend visual and language instructions. These models outperform the current open-source vision-language models across various evaluation benchmarks and support text recognition and visual grounding in both Chinese and English languages. Moreover, these models enable multi-image conversations and storytelling. Further details can be found in~\citet{qwenvl}.
\end{enumerate}

Now, we officially open-source the 14B-parameter and 7B-parameter base pretrained models \qwen{} and aligned chat models \qwenchat{}\footnote{GitHub: \url{https://github.com/QwenLM/Qwen}.}. 
This release aims at providing more comprehensive and powerful LLMs at developer- or application-friendly scales.

\begin{figure}[tb]
\centering
\includegraphics[width=.9\textwidth]{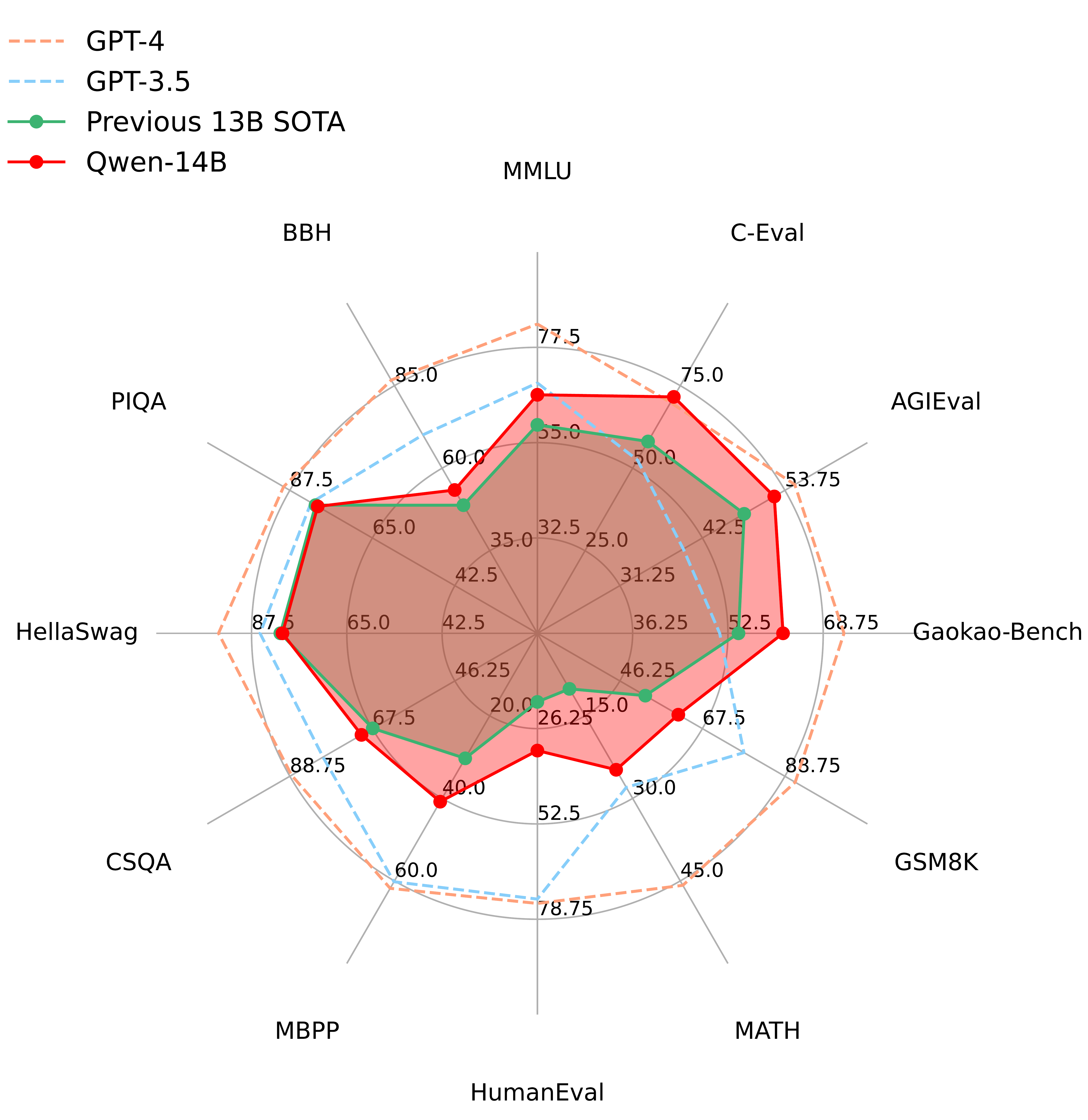}
\caption{\textbf{Performance of GPT-4, GPT-3.5, the previous 13B SOTA, as well as \qwen{}\textsc{-14B}.} We demonstrate the results on \num{12} datasets covering multiple domains, including language understanding, knowledge, reasoning, etc. \qwen{} significantly outperforms the previous SOTA of similar model sizes, but still lag behind both GPT-3.5 and GPT-4.}
\label{fig:results}
\end{figure}

The structure of this report is as follows: Section~\ref{sec:pretraining} describes our approach to pretraining and results of \qwen{}. Section~\ref{sec:alignment} covers our methodology for alignment and reports the results of both automatic evaluation and human evaluation. Additionally, this section describes details about our efforts in building chat models capable of tool use, code interpreter, and agent. In Sections~\ref{sec:coding} and~\ref{sec:math}, we delve into specialized models of coding and math and their performance. Section~\ref{sec:related-work} provides an overview of relevant related work, and Section~\ref{sec:conclusion} concludes this paper and points out our future work.

\section{Pretraining}
\label{sec:pretraining}

The pretraining stage involves learning vast amount of data to acquire a comprehensive understanding of the world and its various complexities. This includes not only basic language capabilities but also advanced skills such as arithmetic, coding, and logical reasoning. 
In this section, we introduce the data, the model design and scaling, as well as the comprehensive evaluation results on benchmark datasets. 

\subsection{Data}
The size of data has proven to be a crucial factor in developing a robust large language model, as highlighted in previous research~\citep{chinchilla, llama2}. To create an effective pretraining dataset, it is essential to ensure that the data are diverse and cover a wide range of types, domains, and tasks. Our dataset is designed to meet these requirements and includes public web documents, encyclopedia, books, codes, etc. Additionally, our dataset is multilingual, with a significant portion of the data being in English and Chinese. 

To ensure the quality of our pretraining data, we have developed a comprehensive data preprocessing procedure. For public web data, we extract text from HTML and use language identification tools to determine the language. To increase the diversity of our data, we employ deduplication techniques, including exact-match deduplication after normalization and fuzzy deduplication using MinHash and LSH algorithms.
To filter out low-quality data, we employ a combination of rule-based and machine-learning-based methods. Specifically, we use multiple models to score the content, including language models, text-quality scoring models, and models for identifying potentially offensive or inappropriate content. We also manually sample texts from various sources and review them to ensure their quality.
To further enhance the quality of our data, we selectively up-sample data from certain sources, to ensure that our models are trained on a diverse range of high-quality content. 
In recent studies~\citep{glm-130b,ext5,t5}, it has been demonstrated that pretraining language models with multi-task instructions can enhance their zero-shot and few-shot performance. To further enhance the performance of our model, we have incorporated high-quality instruction data into our pretraining process.
To safeguard the integrity of our benchmark assessment, we have adopted a similar approach as~\citet{gpt3} and meticulously eliminated any instruction samples that exhibit a 13-gram overlap with any data present in the test sets utilized in our evaluation. 
Given the large number of downstream tasks, it is not feasible to repeat this filtering process for all tasks. Instead, we have made sure that the instruction data for the reported tasks have undergone our filtering process to ensure their accuracy and reliability.
Finally, we have built a dataset of up to \num{3} trillion tokens.

\subsection{Tokenization} 
The design of vocabulary significantly impacts the training efficiency and the downstream task performance.
In this study, we utilize byte pair encoding (BPE) as our tokenization method, following GPT-3.5 and GPT-4. We start with the open-source fast BPE tokenizer, tiktoken~\citep{tiktoken}, and select the vocabulary \textrm{cl100k} base as our starting point.
To enhance the performance of our model on multilingual downstream tasks, particularly in Chinese, we augment the vocabulary with commonly used Chinese characters and words, as well as those in other languages. 
Also, following \citet{llama, llama2}, we have split numbers into single digits. 
The final vocabulary size is approximately \num{152}K.

\begin{figure}[tbp]
\includegraphics[width= 1\textwidth]{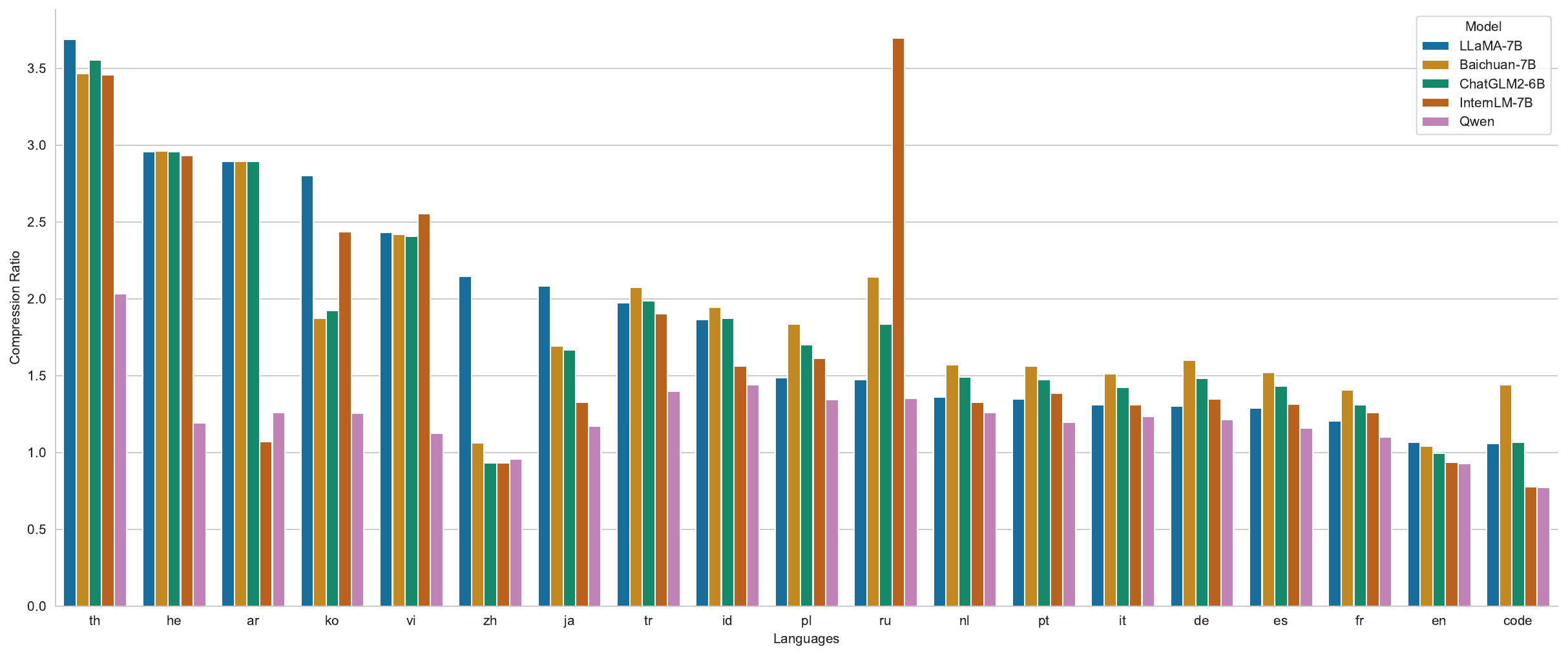}
\caption{\textbf{Encoding compression rates of different models.} We randomly selected \num{1} million document corpora of each language to test and compare the encoding compression rates of different models (with XLM-R~\citep{xlmr}, which supports \num{100} languages, as the base value \num{1}, not shown in the figure). As can be seen, while ensuring the efficient decoding of Chinese, English, and code, \qwen{} also achieves a high compression rate for many other languages (such as th, he, ar, ko, vi, ja, tr, id, pl, ru, nl, pt, it, de, es, fr, etc.), equipping the model with strong scalability as well as high training and inference efficiency in these languages.}
\label{fig:tokenizer}
\end{figure}

The performance of the \qwen{} tokenizer in terms of compression is depicted in Figure~\ref{fig:tokenizer}. In this comparison, we have evaluated \qwen{} against several other tokenizers, including XLM-R~\citep{xlmr}, LLaMA~\citep{llama}, Baichuan~\citep{baichuan7b}, and InternLM~\citep{internlm}. Our findings reveal that \qwen{} achieves higher compression efficiency than its competitors in most languages. This implies that the cost of serving can be significantly reduced since a smaller number of tokens from \qwen{} can convey more information than its competitors.
Furthermore, we have conducted preliminary experiments to ensure that scaling the vocabulary size of \qwen{} does not negatively impact the downstream performance of the pretrained model. Despite the increase in vocabulary size, our experiments have shown that \qwen{} maintains its performance levels in downstream evaluation.

\subsection{Architecture}

\qwen{} is designed using a modified version of the Transformer architecture. Specifically, we have adopted the recent open-source approach of training large language models, LLaMA~\citep{llama}, which is widely regarded as the top open-source LLM. Our modifications to the architecture include:
\begin{itemize}
    \item \textbf{Embedding and output projection}. Based on preliminary experimental findings, we have opted for the untied embedding approach instead of tying the weights of input embedding and output projection. This decision was made in order to achieve better performance with the price of memory costs.
    \item \textbf{Positional embedding}. We have chosen RoPE (Rotary Positional Embedding)~\citep{rope} as our preferred option for incorporating positional information into our model. RoPE has been widely adopted and has demonstrated success in contemporary large language models, notably PaLM~\citep{palm, palm2} and LLaMA~\citep{llama, llama2}. In particular, we have opted to use FP32 precision for the inverse frequency matrix, rather than BF16 or FP16, in order to prioritize model performance and achieve higher accuracy.
    \item \textbf{Bias}. For most layers, we remove biases following \citet{palm}, but we add biases in the QKV layer of attention to enhance the extrapolation ability of the model~\citep{qkv_bias}.
    \item \textbf{Pre-Norm \& RMSNorm}. In modern Transformer models, pre-normalization is the most widely used approach, which has been shown to improve training stability compared to post-normalization. Recent research has suggested alternative methods for better training stability, which we plan to explore in future versions of our model. Additionally, we have replaced the traditional layer normalization technique described in~\citep{layer_norm} with RMSNorm~\citep{rmsnorm}. This change has resulted in equivalent performance while also improving efficiency.
    \item \textbf{Activation function}. We have selected SwiGLU~\citep{noamglu} as our activation function, a combination of Swish~\citep{swish} and Gated Linear Unit~\citep{glu}. Our initial experiments have shown that activation functions based on GLU generally outperform other baseline options, such as GeLU~\citep{gelu}. As is common practice in previous research, we have reduced the dimension of the feed-forward network (FFN) from \num{4} times the hidden size to $\frac{8}{3}$ of the hidden size.
\end{itemize}

\begin{table}[tbp]
\centering
\small
\caption{\textbf{Model sizes, architectures, and optimization hyper-parameters.}}
\label{tab:model-size}
\begin{tabular}{@{}ccccccc@{}}
\toprule
{\# of Params} & Hidden size & Heads & Layers & Learning rate & Batch size & Training tokens \\
\midrule
1.8B     & 2048        & 16      & 24       & $3.0 \times 10 ^ {-4} $     & 4M         & 2.2T            \\
7B       & 4096        & 32      & 32       & $3.0 \times 10 ^ {-4} $      & 4M         & 2.4T            \\
14B      & 5120        & 40      & 40       & $3.0 \times 10 ^ {-4} $      & 4M         & 3.0T             \\\bottomrule
\end{tabular}
\end{table}

\subsection{Training}\label{sec:training}
To train \qwen{}, we follow the standard approach of autoregressive language modeling, as described in~\citet{gpt}. This involves training the model to predict the next token based on the context provided by the previous tokens.
We train models with context lengths of \num{2048}. To create batches of data, we shuffle and merge the documents, and then truncate them to the specified context lengths. 
To improve computational efficiency and reduce memory usage, we employ Flash Attention in the attention modules~\citep{flashattn}.
We adopt the standard optimizer AdamW~\citep{adam,adamw} for pretraining optimization. We set the hyperparameters $\beta_1=0.9$, $\beta_2=0.95$, and $\epsilon=10^{-8}$. We use a cosine learning rate schedule with a specified peak learning rate for each model size.  The learning rate is decayed to a minimum learning rate of $10\%$ of the peak learning rate. 
All the models are trained with BFloat16 mixed precision for training stability. 

\subsection{Context Length Extension} 

Transformer models have a significant limitation in terms of the context length for their attention mechanism. As the context length increases, the quadratic-complexity computation leads to a drastic increase in both computation and memory costs. In this work, we have implemented simple training-free techniques that are solely applied during inference to extend the context length of the model. One of the key techniques we have used is NTK-aware interpolation~\citep{ntk}. Unlike position interpolation (PI)~\citep{position_interpolation} which scales each dimension of RoPE equally, NTK-aware interpolation adjusts the base of RoPE to prevent the loss of high-frequency information in a training-free manner. To further improve performance, we have also implemented a trivial extension called dynamic NTK-aware interpolation, which is later formally discussed in~\citep{yarn}. It dynamically changes the scale by chunks, avoiding severe performance degradation. These techniques allow us to effectively extend the context length of Transformer models without compromising their computational efficiency or accuracy.

\qwen{} additionally incorporates two attention mechanisms: LogN-Scaling~\citep{logn_attn,logn_su} and window attention~\citep{window_attn}. LogN-Scaling rescales the dot product of the query and value by a factor that depends on the ratio of the context length to the training length, ensuring that the entropy of the attention value remains stable as the context length grows.
Window attention restricts the attention to a limited context window, preventing the model from attending to tokens that are too far away.

We also observed that the long-context modeling ability of our model varies across layers, with lower layers being more sensitive in context length extension compared to the higher layers.
To leverage this observation, we assign different window sizes to each layer, using shorter windows for lower layers and longer windows for higher layers.

\begin{table}[tbp]
\centering
\small
\caption{\textbf{Overall performance on widely-used benchmarks compared to open-source base models}. Our largest \qwen{} model with 14 billion parameters outperforms previous 13B SoTA models on all datasets.}
\label{tab:res-overall}
\begin{tabular}{@{}lccccccccc@{}}
\arrayrulecolor{black}\toprule
\textbf{Model} &  \textbf{Params}   & \textbf{MMLU}      & \textbf{C-Eval}  & \textbf{GSM8K} &  \textbf{MATH}   & \textbf{HumanEval}  & \textbf{MBPP}     & \textbf{BBH}      \\ 
                 &  & 5-shot    & 5-shot      &  8-shot & 4-shot   & 0-shot     & 3-shot   & 3-shot   \\ \midrule
\multirow{2}{*}{MPT} & 7B & 30.8 & 23.5 & 9.1 & 3.0 & 18.3 & 22.8 & 35.6  \\
 & \textcolor{gray}{30B} & \textcolor{gray}{47.9} & \textcolor{gray}{-} & \textcolor{gray}{15.2} & \textcolor{gray}{3.1} & \textcolor{gray}{25.0} & \textcolor{gray}{32.8} & \textcolor{gray}{38.0}  \\
 \arrayrulecolor{lightgray}\midrule
\multirow{2}{*}{Falcon} & 7B & 27.8 & - & 6.8 & 2.3 & - & 11.2 & 28.0  \\
 & \textcolor{gray}{40B} & \textcolor{gray}{57.0} & \textcolor{gray}{-}  & \textcolor{gray}{19.6} & \textcolor{gray}{5.5} &  \textcolor{gray}{-} & \textcolor{gray}{29.8} & \textcolor{gray}{37.1}  \\
 \arrayrulecolor{lightgray}\midrule
ChatGLM2 & 6B     & 47.9      & 51.7   & 32.4  & 6.5    & -        & -     & 33.7        \\ 
\arrayrulecolor{lightgray}\midrule
\multirow{2}{*}{InternLM} & 7B     & 51.0      & 53.4      & 31.2 & 6.3      & 10.4       & 14.0     & 37.0       \\ 
&  \textcolor{gray}{20B}     & \textcolor{gray}{62.1}      & \textcolor{gray}{58.8}      & \textcolor{gray}{52.6}   & \textcolor{gray}{7.9}   & \textcolor{gray}{25.6}       & \textcolor{gray}{35.6}     & \textcolor{gray}{52.5}       \\ \arrayrulecolor{lightgray}\midrule
\multirow{2}{*}{Baichuan2} & 7B    & 54.7      & 56.3    & 24.6  & 5.6     & 18.3       & 24.2     & 41.6        \\
  & 13B & 59.5      & 59.0     & 52.8 & 10.1      & 17.1       & 30.2     & 49.0        \\ \midrule
\multirow{4}{*}{LLaMA} & 7B       & 35.6      & 27.3    & 11.0 & 2.9     & 12.8       & 17.7     & 33.5        \\
     & 13B      & 47.7      & 31.8      & 20.3 & 4.2     & 15.8       & 22.0     & 37.9        \\
 & \textcolor{gray}{33B}      & \textcolor{gray}{58.7}      & \textcolor{gray}{37.5}       & \textcolor{gray}{42.3}  & \textcolor{gray}{7.1}    & \textcolor{gray}{21.7}       & \textcolor{gray}{30.2}     & \textcolor{gray}{50.0}           \\
 & \textcolor{gray}{65B}      & \textcolor{gray}{63.7}      & \textcolor{gray}{40.4}    & \textcolor{gray}{54.4}  & \textcolor{gray}{10.6}    & \textcolor{gray}{23.7}       &  \textcolor{gray}{37.7}     & \textcolor{gray}{58.4}          \\
\midrule
\multirow{4}{*}{\llamatwo{}} & 7B       & 46.8      & 32.5      & 16.7   & 3.3   & 12.8       & 20.8     & 38.2      \\
     & 13B      & 55.0      & 41.4     & 29.6  & 5.0    & 18.9       & 30.3     & 45.6       \\
 & \textcolor{gray}{34B}      & \textcolor{gray}{62.6}      & \textcolor{gray}{-}     & \textcolor{gray}{42.2}  & \textcolor{gray}{6.2}    & \textcolor{gray}{22.6}       & \textcolor{gray}{33.0}     & \textcolor{gray}{44.1}          \\
 & \textcolor{gray}{70B}      & \textcolor{gray}{69.8}      & \textcolor{gray}{50.1}    & \textcolor{gray}{63.3}  & \textcolor{gray}{13.5}    & \textcolor{gray}{29.9}       &  \textcolor{gray}{45.0}     & \textcolor{gray}{ 64.9}           \\ \midrule
StableBeluga2 & \textcolor{gray}{70B}    & \textcolor{gray}{68.6}   & \textcolor{gray}{51.4}   & \textcolor{gray}{69.6}   & \textcolor{gray}{14.6}       & \textcolor{gray}{28.0}       & \textcolor{gray}{11.4}     & \textcolor{gray}{69.3}        \\ \midrule
\multirow{3}{*}{\qwen{}} & 1.8B   & 44.6      & 54.7      & 21.2  & 5.6    & 17.1       & 14.8        & 28.2    \\
 & 7B     & 58.2      & 63.5    & 51.7   & 11.6   & 29.9       & 31.6     & 45.0       \\
 & 14B    &  \bf 66.3  & \bf 72.1  & \bf 61.3 & \bf 24.8 & \bf 32.3   & \bf 40.8 &  \bf 53.4   \\
\arrayrulecolor{black}\bottomrule
\end{tabular}
\end{table}

\subsection{Experimental Results}
To evaluate the zero-shot and few-shot learning capabilities of our models, we conduct a thorough benchmark assessment using a series of datasets. We compare \qwen{} with the most recent open-source base models, including LLaMA~\citep{llama}, \llamatwo{}~\citep{llama2}, MPT~\citep{mpt}, Falcon~\citep{falcon}, Baichuan2~\citep{baichuan2}, ChatGLM2~\citep{chatglm2}, InternLM~\citep{internlm}, XVERSE~\citep{xverse}, and StableBeluga2~\citep{StableBeluga2}. Our evaluation covers a total of 7 popular benchmarks, which are MMLU (5-shot)~\citep{mmlu}, C-Eval (5-shot)~\citep{ceval}, GSM8K (8-shot)~\citep{gsm8k}, MATH (4-shot)~\citep{hendrycks2021measuring}, HumanEval (0-shot)~\citep{codex}, MBPP (0-shot)~\citep{mbpp}, and BBH (Big Bench Hard) (3 shot)~\citep{bbh}. We aim to provide a comprehensive summary of the overall performance of our models across these benchmarks.

In this evaluation, we focus on the base language models without alignment and collect the baselines' best scores from their official results and OpenCompass~\citep{opencompass}.
The results are presented in Table~\ref{tab:res-overall}.

Our experimental results demonstrate that the three \qwen{} models exhibit exceptional performance across all downstream tasks. It is worth noting that even the larger models, such as LLaMA2-70B, are outperformed by \qwen{}\textsc{-14B} in \num{3} tasks. 
\qwen{}\textsc{-7B} also performs admirably, surpassing LLaMA2-13B and achieving comparable results to Baichuan2-13B. Notably, despite having a relatively small number of parameters, \qwen{}\textsc{-1.8B} is capable of competitive performance on certain tasks and even outperforms larger models in some instances.
The findings highlight the impressive capabilities of the \qwen{} models, particularly \qwen{}\textsc{-14B}, and suggest that smaller models, such as \qwen{}\textsc{-1.8B}, can still achieve strong performance in certain applications.

\begin{table}[t]
\centering
\small
\caption{\textbf{Results of \qwen{} on long-context inference using various techniques.} Our experimental findings reveal that the application of our crucial techniques enables the model to consistently achieve low perplexity as the context length increases. This suggests that these techniques play a significant role in enhancing the model's ability to comprehend and generate lengthy texts.}
\label{tab:res-long}
\begin{tabular}{@{}lccccc@{}}
\toprule
\multirow{2}{*}[-\aboverulesep]{\textbf{Model}}              & \multicolumn{5}{c}{\textbf{Sequence Length}}    \\ \cmidrule(l){2-6}
 & \multicolumn{1}{c}{1024} & \multicolumn{1}{c}{2048} & \multicolumn{1}{c}{4096} & \multicolumn{1}{c}{8192} & \multicolumn{1}{c}{16384} \\ \midrule
\qwen{}\textsc{-7B}                             & 4.23 & 3.78 & 39.35 & 469.81 & 2645.09 \\
+ dynamic\_ntk                      & 4.23 & 3.78 & 3.59  & 3.66   & 5.71    \\
+ dynamic\_ntk + logn               & 4.23 & 3.78 & 3.58  & 3.56   & 4.62    \\
+ dynamic\_ntk + logn + window\_attn & 4.23 & 3.78 & 3.58  & 3.49   & 4.32   \\ \midrule
\qwen{}\textsc{-14B} & - & 3.46 & 22.79 & 334.65 & 3168.35 \\
+ dynamic\_ntk + logn + window\_attn & - & 3.46 & 3.29 & 3.18   & 3.42   \\
\bottomrule
\end{tabular}
\end{table}

To evaluate the effectiveness of context length extension, Table~\ref{tab:res-long} presents the test results on arXiv\footnote{The dataset contains academic papers from \url{https://arxiv.org}.} in terms of perplexity (PPL). 
These results demonstrate that by combining NTK-aware interpolation, LogN-Scaling, and layer-wise window assignment, we can effectively maintain the performance of our models in the context of over \num{8192} tokens. 

\section{Alignment}
\label{sec:alignment}

Pretrained large language models have been found to be not aligned with human behavior, making them unsuitable for serving as AI assistants in most cases. Recent research has shown that the use of alignment techniques, such as supervised finetuning (SFT) and reinforcement learning from human feedback (RLHF), can significantly improve the ability of language models to engage in natural conversation. In this section, we will delve into the details of how \qwen{} models have been trained using SFT and RLHF, and evaluate their performance in the context of chat-based assistance.

\subsection{Supervised Finetuning}
To gain an understanding of human behavior, the initial step is to carry out SFT, which finetunes a pretrained LLM on chat-style data, including both queries and responses. In the following sections, we will delve into the details of data construction and training methods.

\subsubsection{Data}

To enhance the capabilities of our supervised finetuning datasets, we have annotated conversations in multiple styles. While conventional datasets~\citep{flan} contain a vast amount of data prompted with questions, instructions, and answers in natural language, our approach takes it a step further by annotating human-style conversations. This practice, inspired by~\citet{instructgpt}, aims at improving the model's helpfulness by focusing on natural language generation for diverse tasks.
To ensure the model's ability to generalize to a wide range of scenarios, we specifically excluded data formatted in prompt templates that could potentially limit its capabilities. Furthermore, we have prioritized the safety of the language model by annotating data related to safety concerns such as violence, bias, and pornography. 

In addition to data quality, we have observed that the training method can significantly impact the final performance of the model. To achieve this, we utilized the ChatML-style format~\citep{chatml}, which is a versatile meta language capable of describing both the metadata (such as roles) and the content of a turn. This format enables the model to effectively distinguish between various types of information, including system setup, user inputs, and assistant outputs, among others. By leveraging this approach, we can enhance the model's ability to accurately process and analyze complex conversational data.

\subsubsection{Training}
Consistent with pretraining, we also apply next-token prediction as the training task for SFT. We apply the loss masks for the system and user inputs. More details are demonstrated in Section~\ref{appendix:data_format}. 

The model's training process utilizes the AdamW optimizer, with the following hyperparameters: $\beta_1$ set to \num{0.9}, $\beta_2$ set to \num{0.95},  and $\epsilon$ set to $10^{-8}$.
The sequence length is limited to \num{2048}, and the batch size is \num{128}.
The model undergoes a total of \num{4000} steps,  with the learning rate gradually increased over the first \num{1430} steps, reaching a peak of $2 \times 10^{-6}$.
To prevent overfitting, weight decay is applied with a value of \num{0.1}, dropout is set to \num{0.1}, and gradient clipping is enforced with a limit of \num{1.0}.

\subsection{Reinforcement Learning from Human Feedback}

While SFT has proven to be effective, we acknowledge that its generalization and creativity capabilities may be limited, and it is prone to overfitting. To address this issue, we have implemented Reinforcement Learning from Human Feedback (RLHF) to further align SFT models with human preferences, following the approaches of~\citet{instructgpt, rlhf}. This process involves training a reward model and using Proximal Policy Optimization (PPO)~\citep{ppo} to conduct policy training.

\subsubsection{Reward Model}
To create a successful reward model, like building a large language model (LLM), it is crucial to first undergo pretraining and then finetuning. This pretraining process, also known as preference model pretraining (PMP)~\citep{PMP}, necessitates a vast dataset of comparison data. This dataset consists of sample pairs, each containing two distinct responses for a single query and their corresponding preferences. Similarly, finetuning is also conducted on this type of comparison data, but with a higher quality due to the presence of quality annotations.

During the fine-tuning phase, we gather a variety of prompts and adjust the reward model based on human feedback for responses from the \qwen{} models. To ensure the diversity and complexity of user prompts are properly taken into account, we have created a classification system with around \num{6600} detailed tags and implemented a balanced sampling algorithm that considers both diversity and complexity when selecting prompts for annotation by the reward model~\citep{instag}. To generate a wide range of responses, we have utilized \qwen{} models of different sizes and sampling strategies, as diverse responses can help reduce annotation difficulties and enhance the performance of the reward model. These responses are then evaluated by annotators following a standard annotation guideline, and comparison pairs are formed based on their scores. 

In creating the reward model, we utilize the same-sized pre-trained language model \qwen{} to initiate the process. It is important to mention that we have incorporated a pooling layer into the original \qwen{} model to extract the reward for a sentence based on a specific end token.
The learning rate for this process has been set to a constant value of $3 \times 10^{-6}$,  and the batch size is \num{64}. Additionally, the sequence length is set to \num{2048}, and the training process lasts for a single epoch.

We adopted the accuracy on the test dataset as an important but not exclusive evaluation metric for the reward model. In Table~\ref{tab:res-rm}, we report the test pairwise accuracy of PMP and reward models on diverse human preference benchmark datasets~\citep{PMP,learn-summary, pmlr-v162-ethayarajh22a, lightman2023lets}. Specifically, \qwen{} Helpful-base and \qwen{} Helpful-online are our proprietary datasets. The responses in \qwen{} Helpful-base are generated from \qwen{} without RLHF, whereas \qwen{} Helpful-online includes responses from \qwen{} with RLHF. The results show that the PMP model demonstrates high generalization capabilities on out-of-distribution data, and the reward model demonstrates significant improvement on our \qwen{} reward datasets.

\begin{table}[tbp]
\centering
\small
\caption{Test Accuracy of \qwen{} preference
model pretraining (PMP) and reward model (RM) on diverse human preference benchmark datasets.}
\label{tab:res-rm}
\setlength{\tabcolsep}{5pt}
\begin{tabular}{@{}lccccccc@{}}
\toprule
 \multirow{2}{*}{Dataset}  & \qwen{}   & \qwen{} &  Anthropic  & Anthropic  & OpenAI  & Stanford & OpenAI  \\ 
  &  Helpful-base  & Helpful-online &  Helpful-base &  Helpful-online &  Summ. &  SHP & PRM800K \\\midrule
PMP    &     62.68          &         61.62              &       76.52                &      65.43    &  69.60       &     60.05    &      70.59          \\
RM     &     74.78         &          69.71             &        73.98               &       64.57    &  69.99       &     60.10    &      70.52   
\\\bottomrule
\end{tabular}
\label{tb:rm_acc}
\end{table}

\subsubsection{Reinforcement Learning}

Our Proximal Policy Optimization (PPO) process involves four models: the policy model, value model, reference model, and reward model. Before starting the PPO procedure, we pause the policy model's updates and focus solely on updating the value model for \num{50} steps. This approach ensures that the value model can adapt to different reward models effectively.

During the PPO operation, we use a strategy of sampling two responses for each query simultaneously. This strategy has proven to be more effective based on our internal benchmarking evaluations. We set the KL divergence coefficient to \num{0.04} and normalize the reward based on the running mean. 

The policy and value models have learning rates of $1 \times 10^{-6}$ and $5 \times 10^{-6}$, respectively. To enhance training stability, we utilize value loss clipping with a clip value of \num{0.15}. For inference, the policy top-p is set to \num{0.9}. Our findings indicate that although the entropy is slightly lower than when top-p is set to \num{1.0}, there is a faster increase in reward, ultimately resulting in consistently higher evaluation rewards under similar conditions.

Additionally, we have implemented a pretrained gradient to mitigate the alignment tax. 
Empirical findings indicate that, with this specific reward model, the KL penalty is adequately robust to counteract the alignment tax in benchmarks that are not strictly code or math in nature, such as those that test common sense knowledge and reading comprehension. It is imperative to utilize a significantly larger volume of the pretrained data in comparison to the PPO data to ensure the effectiveness of the pretrained gradient. Additionally, our empirical study suggests that an overly large value for this coefficient can considerably impede the alignment to the reward model, eventually compromising the ultimate alignment, while an overly small value would only have a marginal effect on alignment tax reduction.

\subsection{Automatic and Human Evaluation of Aligned Models}

To showcase the effectiveness of our aligned models, we conduct a comparison with other aligned models on well-established benchmarks, including MMLU~\citep{mmlu}, C-Eval~\citep{ceval}, GSM8K~\citep{gsm8k}, HumanEval~\citep{codex}, and BBH~\citep{bbh}. 
Besides the widely used few-shot setting, we test our aligned models in the zero-shot setting to demonstrate how well the models follow instructions. The prompt in a zero-shot setting consists of an instruction and a question without any previous examples in the context. The results of the baselines are collected from their official reports and OpenCompass~\citep{opencompass}.

The results in Table~\ref{tab:res-overall-chat} demonstrate the effectiveness of our aligned models in understanding human instructions and generating appropriate responses. \qwen{}-14B-Chat outperforms all other models except ChatGPT~\citep{chatgpt} and \llamatwochat{}\textsc{-70B}~\citep{llama2} in all datasets, including MMLU~\citep{mmlu}, C-Eval~\citep{ceval}, GSM8K~\citep{gsm8k}, HumanEval~\citep{codex}, and BBH~\citep{bbh}. In particular, \qwen{}'s performance in HumanEval, which measures the quality of generated codes, is significantly higher than that of other open-source models. 

Moreover, \qwen{}'s performance is consistently better than that of open-source models of similar size, such as LLaMA2~\citep{llama2}, ChatGLM2~\citep{chatglm2}, InternLM~\citep{internlm}, and Baichuan2~\citep{baichuan2}. This suggests that our alignment approach, which involves fine-tuning the model on a large dataset of human conversations, has been effective in improving the model's ability to understand and generate human-like language.

\begin{table}[tbp]
\centering
\small
\caption{\textbf{Performance of aligned models on widely-used benchmarks.} We report both zero-shot and few-shot performance of the models.}
\label{tab:res-overall-chat}
\resizebox{0.95\columnwidth}{!}{
\begin{tabular}{@{}lcc@{ / }cc@{ / }cc@{ / }ccc@{ / }c@{}}
\toprule
\textbf{Model} &  \textbf{Params}          & \multicolumn{2}{c}{\bf MMLU}      & \multicolumn{2}{c}{\bf C-Eval}    & \multicolumn{2}{c}{\bf GSM8K}     & \bf HumanEval      & \multicolumn{2}{c}{\bf BBH}    \\ 
&    & 0-shot & 5-shot    &  0-shot & 5-shot   & 0-shot & 8-shot     & 0-shot   & 0-shot & 3-shot  \\ \midrule
\multicolumn{11}{c}{\textit{Proprietary models}}\\
\arrayrulecolor{black}\midrule
GPT-3.5 & -   & - & 69.1     & - & 52.5   & -   & 78.2      & 73.2      & -   & 70.1       \\
GPT-4 & - & - & \textbf{83.0}   & - & \textbf{69.9}   & -  & \textbf{91.4}     & \textbf{86.6}      & -   & \textbf{86.7}       \\
\arrayrulecolor{black}\midrule
\multicolumn{11}{c}{\textit{Open-source models}}\\
\arrayrulecolor{black}\midrule
\multirow{1}{*}{ChatGLM2}   & 6B & 45.5 &  46.0    & 50.1 & 52.6  & -   & 28.8      & 11.0    & -     & 32.7     \\
 \arrayrulecolor{lightgray}\midrule
InternLM-Chat  & 7B  & - & 51.1   & -   & 53.6   & -   & 33.0      & 14.6       & -  & 32.5     \\ \midrule
\multirow{2}{*}{Baichuan2-Chat}  & 7B  & - & 52.9  &  - &   55.6    & -  & 32.8      & 13.4       & -    & 35.8      \\
  & 13B & - & 57.3    & -  & 56.7     & - & 55.3      & 17.7      & -    & 49.9    \\ \midrule
\multirow{3}{*}{\llamatwochat{}} & 7B  & - & 46.2      & - & 31.9     & - & 26.3      & 12.2      & -    & 35.6  \\ 
& 13B  &  - &  54.6      & - & 36.2    & -  & 37.1      & 18.9     & -    & 40.1    \\
& \textcolor{gray}{70B}   & \textcolor{gray}{-} & \textcolor{gray}{63.8}     & \textcolor{gray}{-} & \textcolor{gray}{44.3}   & \textcolor{gray}{-}   & \textcolor{gray}{59.3}      & \textcolor{gray}{32.3}      & \textcolor{gray}{-}    & \textcolor{gray}{60.8}       \\
  \arrayrulecolor{black}\midrule
\multirow{3}{*}{\qwenchat{}}  & 1.8B & 42.4 & 43.9      & 50.7 & 50.3      & 27.8 & 19.5   & 14.6         & 27.1 & 25.0     \\
  & 7B   & 55.8 & 57.0      & 59.7 & 59.3      & 50.3 & 54.1      & 37.2           & 39.6 & 46.7     \\
 & 14B   & 64.6 & \bf 66.5      & 69.8 & \bf 71.7      & \bf 60.1 & 59.3  & \bf 43.9           & 46.9 & \bf 58.7    \\ \bottomrule
\end{tabular}}
\end{table}

\begin{figure}[htbp]
\includegraphics[width= 1\textwidth]{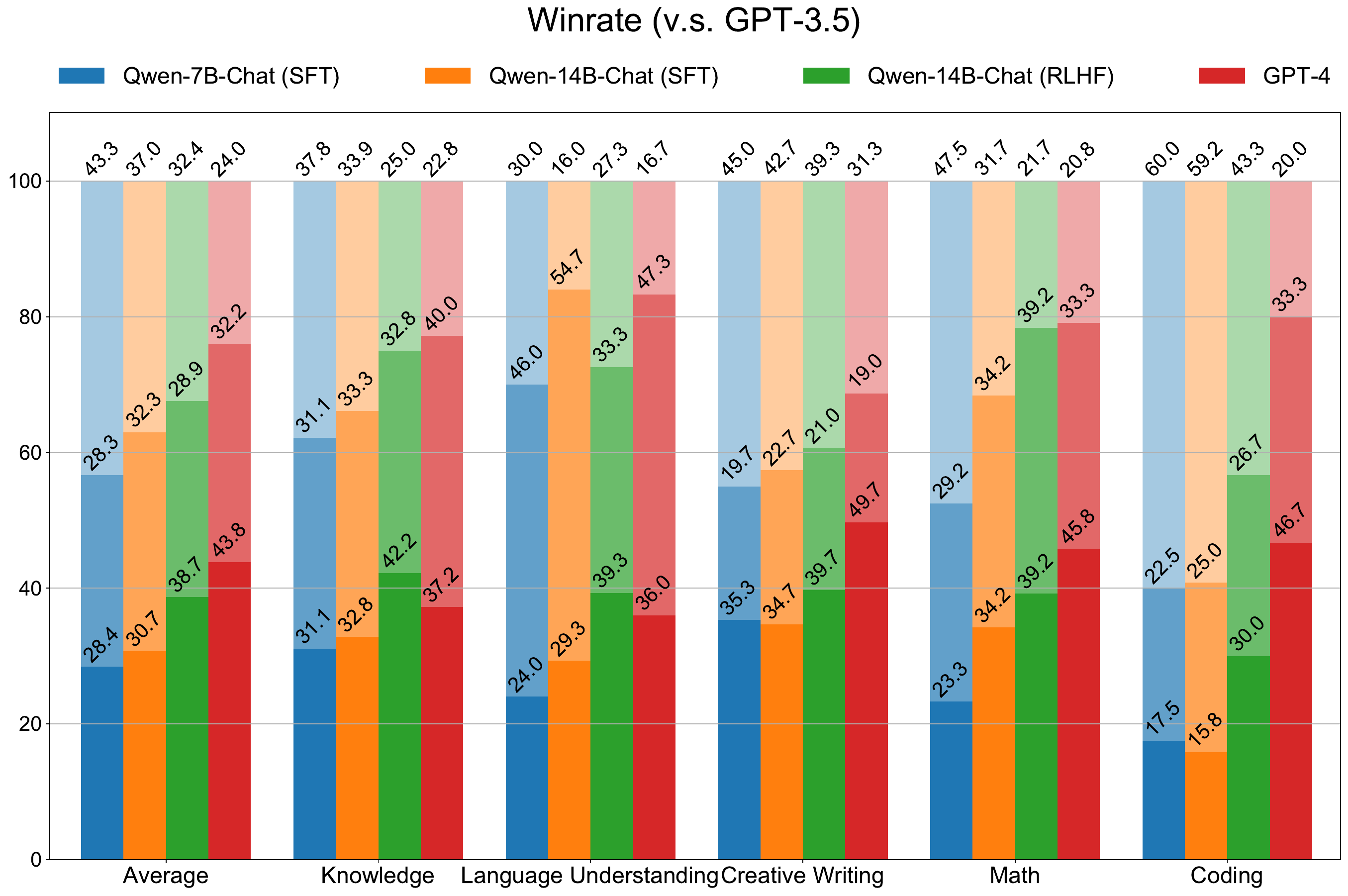}
\caption{\textbf{Results of the human evaluation for chat models.} We compare Qwen-7B (SFT), Qwen-14B (SFT), Qwen-14B (RLHF), as well as GPT-4 against GPT-3.5. Each bar segment represents the percentage of wins, ties, and losses, from bottom to top. On average, the RLHF model outperforms the SFT model. The dataset consists of 300 Chinese instructions.}
\label{fig:humaneval-winrate}
\end{figure}

Despite this, we have reservations about the ability of traditional benchmark evaluation to accurately measure the performance and potential of chat models trained with alignment techniques in today's landscape. The results mentioned earlier provide some evidence of our competitive standing, but we believe that it is crucial to develop new evaluation methods specifically tailored to aligned models.

We believe that human evaluation is crucial, which is why we have created a carefully curated dataset for this purpose. Our process involved collecting \num{300} instructions in Chinese that covered a wide range of topics, including knowledge, language understanding, creative writing, coding, and mathematics. 
To evaluate the performance of different models, we chose the SFT version of \qwenchat{}-7B and the SFT and RLHF versions of \qwenchat{}-14B, and added two strong baselines, GPT-3.5 and GPT-4\footnote{To obtain the results from the models, we use the OpenAI APIs of GPT-3.5-turbo-0613 and GPT-4-0613.}, for comparison. For each instruction, we asked three annotators to rank the model responses by the overall score of helpfulness, informativeness, validity, and other relevant factors.
Our dataset and evaluation methodology provides a comprehensive and rigorous assessment of the capabilities of different language models in various domains.

Figure~\ref{fig:humaneval-winrate} illustrates the win rates of the various models. For each model, we report the percentage of wins, ties, and losses against GPT-3.5, with the segments of each bar from bottom to top representing these statistics.
The experimental results clearly demonstrate that the RLHF model outperforms the SFT models by significant margins, indicating that RLHF can encourage the model to generate responses that are more preferred by humans. In terms of overall performance, we find that the RLHF model significantly outperforms the SFT models, falling behind GPT-4. 
This indicates the effectiveness of RLHF for aligning to human preference. 
To provide a more comprehensive understanding of the models' performance, we include a case study with examples from different models in Appendix~\ref{app:cases}.
Nonetheless, it remains difficult to accurately capture the gap between our models and the proprietary models. As such, a more extensive and rigorous assessment is required for the chat models.

\subsection{Tool Use, Code Interpreter, and Agent}\label{sec:tool_use}

\begin{table}[htbp]
\centering
\small
\caption{Performance of \qwen{} on the in-house Chinese benchmark that evaluates its ability to use unseen tools via ReAct prompting.}
\label{tab:res-react-chat}
\begin{tabular}{@{}lcccc@{}}
\toprule
\textbf{Model}   & \textbf{Params}     & \textbf{Tool Selection (Acc.↑)} & \textbf{Tool Input (Rouge-L↑)} & \textbf{False Positive Error (\%)↓} \\ \midrule
GPT-4    & -    & 95                   & 90              & 15.0                \\
\arrayrulecolor{lightgray}\midrule
GPT-3.5  & -    & 85                  & 88                  & 75.0                \\
\arrayrulecolor{black}\midrule
\multirow{3}{*}{\qwenchat{}} & 1.8B  & 92 & 89 & 19.3 \\
 & 7B  & \textbf{98}               & 91                  & \phantom{0}7.3        \\
 & 14B  & \textbf{98} & \textbf{93} & \textbf{\phantom{0}2.4} \\
\bottomrule
\end{tabular}
\end{table}

\begin{table}[htbp]
\centering
\small
\caption{The proportion of code generated by \qwen{} that is executable on the in-house evaluation benchmark for Code Interpreter. This benchmark examines \qwen{}'s coding proficiency in math problem solving, data visualization, and general purposes. \codellama{} underperforms on visualization tasks because it hallucinates non-existent columns solely based on CSV file names (see Figure~\ref{fig:code-interpreter-showcase}).}
\label{tab:res-agent-code-executable-rate}
\begin{tabular}{@{}lccccc@{}}
\toprule
\multirow{2}{*}[-\aboverulesep]{\textbf{Model}} & \multirow{2}{*}[-\aboverulesep]{\textbf{Params}} &  \multicolumn{3}{c}{\textbf{Category}} & \\ \cmidrule(l){3-6}
&  & Math (\%) & Visualization (\%)  & General (\%) & All (\%) \\ \midrule
GPT-4    & - & 91.9 & 85.9 & 82.8 & 86.8 \\
\arrayrulecolor{lightgray}\midrule
GPT-3.5  & - & 89.2 & 65.0 & 74.1 & 72.9 \\
\arrayrulecolor{lightgray}\midrule
\multirow{2}{*}{\llamatwochat{}}  & 7B & 41.9 & 33.1 & 24.1 & 33.6 \\
 & 13B & 50.0 & 40.5 & 48.3 & 44.4 \\
\arrayrulecolor{lightgray}\midrule
\multirow{2}{*}{\codellama{}\textsc{-Instruct}}  & 7B & 85.1 & 54.0 & 70.7 & 65.1 \\
 & 13B & 93.2 & 55.8 & 74.1 & 68.8 \\
\arrayrulecolor{lightgray}\midrule
\multirow{2}{*}{InternLM-Chat} & 7B v1.1 & 78.4 & 44.2 & 62.1 & 56.3 \\
 & 20B & 70.3 & 44.2 & 65.5 & 54.9 \\
\arrayrulecolor{black}\midrule
\multirow{3}{*}{\qwenchat{}} & 1.8B & 33.8 & 30.1 & 8.6 & 26.8 \\
   & 7B          & 82.4 & 64.4 & 67.2 & 70.2 \\
  & 14B          & 89.2 & 84.1 & 65.5 & 81.7 \\ 
\bottomrule
\end{tabular}
\end{table}

\begin{table}[htbp]
\centering
\small
\caption{Correctness of the final response on the in-house evaluation benchmark for Code Interpreter. Visualization-Hard tasks involve planning multiple steps, while Visualization-Easy tasks do not. Visualization-All measures both types of tasks. 
\codellama{} excels in performing Visualization-Easy tasks but tends to underperform in Visualization-Hard tasks, due to its inclination to hallucinate non-existent columns based on the name of a CSV file (see Figure~\ref{fig:code-interpreter-showcase}). 
}
\label{tab:res-agent-code-correctness}
\begin{tabular}{@{}lccccc@{}}
\toprule
\multirow{2}{*}[-\aboverulesep]{\textbf{Model}} & \multirow{2}{*}[-\aboverulesep]{\textbf{Params}} & \multicolumn{4}{c}{\textbf{Category}} \\ \cmidrule(l){3-6}
 & & Math (\%) & Vis.-Hard (\%) & Vis.-Easy (\%) & Vis.-All (\%) \\ \midrule
GPT-4    & - & 82.8 & 66.7 & 60.8 & 63.8 \\
\arrayrulecolor{lightgray}\midrule
GPT-3.5  & - & 47.3 & 33.3 & 55.7 & 44.2 \\
\arrayrulecolor{lightgray}\midrule
\multirow{2}{*}{\llamatwochat{}} & 7B & 3.9 & 14.3 & 39.2 & 26.4 \\
 & 13B & 8.3 & 8.3 & 40.5 & 23.9 \\
\arrayrulecolor{lightgray}\midrule
\multirow{2}{*}{\codellama{}\textsc{-Instruct}}  & 7B & 14.3 & 26.2 & 60.8 & 42.9 \\
  & 13B & 28.2 & 27.4 & 62.0 & 44.2 \\
\arrayrulecolor{lightgray}\midrule
\multirow{2}{*}{InternLM-Chat} & 7B v1.1 & 28.5 & 4.8 & 40.5 & 22.1 \\
 & 20B & 34.6 & 21.4 & 45.6 & 33.1 \\
\arrayrulecolor{black}\midrule
\multirow{3}{*}{\qwenchat{}} & 1.8B & 14.7 & 3.6 & 20.3 & 11.7 \\ 
       & 7B & 41.9 & 40.5 & 54.4 & 47.2 \\
       & 14B & 58.4 & 53.6 & 59.5 & 56.4 \\
\bottomrule
\end{tabular}
\end{table}

\begin{table}[tbp]
\centering
\small
\caption{Results of \qwen{}-Chat on the Hugging Face Agent benchmark.}
\label{tab:res-agent-chat}
\begin{tabular}{@{}llcccc@{}}
\toprule
\multirow{2}{*}[-\aboverulesep]{\textbf{Task}} & \multirow{2}{*}[-\aboverulesep]{\textbf{Model}} & \multirow{2}{*}[-\aboverulesep]{\textbf{Params}} & \multicolumn{3}{c}{\textbf{Metric}} \\ \cmidrule(l){4-6}
 &    &    & Tool Selection $\uparrow $ & Tool Used  $\uparrow $ & Code Correctness $\uparrow $\\
 \arrayrulecolor{black}\midrule
\multirow{7}{*}{Run Mode} & GPT-4  & -   & 100                   & 100                & 97.4                \\
\arrayrulecolor{lightgray}\cmidrule{2-6}
& GPT-3.5  & -   & 95.4                   & 96.3                & 87.0                \\
\arrayrulecolor{lightgray}\cmidrule{2-6}
& Starcoder-Base & 15B & 86.1 & 87.0 & 68.9 \\
\arrayrulecolor{lightgray}\cmidrule{2-6}
& Starcoder & 15B & 87.0 & 88.0 & 68.9 \\
\arrayrulecolor{black}\cmidrule{2-6}
& \multirow{3}{*}{\qwenchat{}} & 1.8B & 85.2 & 84.3 & 61.1 \\ 
&  & 7B & 87.0 & 87.0 & 71.5 \\ 
&  & 14B & 93.5 & 94.4 & 87.0 \\ 
\arrayrulecolor{black}\midrule
\multirow{7}{*}{Chat Mode} & GPT-4  & -   & 97.9 & 97.9 & 98.5               \\
\arrayrulecolor{lightgray}\cmidrule{2-6}
& GPT-3.5  & -   & 97.3 & 96.8 &  89.6                \\
\arrayrulecolor{lightgray}\cmidrule{2-6}
& Starcoder-Base & 15B & 97.9 & 97.9 & 91.1 \\
\arrayrulecolor{lightgray}\cmidrule{2-6}
& Starcoder & 15B & 97.9 & 97.9 & 89.6 \\
\arrayrulecolor{black}\cmidrule{2-6}
& \multirow{3}{*}{\qwenchat{}} & 1.8B & 93.6 & 93.6 & 73.2 \\ 
&  & 7B & 94.7 & 94.7 & 85.1 \\ 
&  & 14B & 97.9 & 97.9 & 95.5 \\ 
\arrayrulecolor{black}\bottomrule
\end{tabular}
\end{table}

The \qwen{} models, which are designed to be versatile, have the remarkable ability to assist with (semi-)automating daily tasks by leveraging their skills in tool-use and planning. As such, they can serve as agents or copilots to help streamline various tasks. We explore \qwen{}'s proficiency in the following areas:
\begin{itemize}
    \item Utilizing unseen tools through ReAct prompting~\citep{yao2022react} (see Table~\ref{tab:res-react-chat}).
    \item Using a Python code interpreter to enhance math reasoning, data analysis, and more (see Table~\ref{tab:res-agent-code-executable-rate} and Table~\ref{tab:res-agent-code-correctness}).
    \item Functioning as an agent that accesses Hugging Face's extensive collection of multimodal models while engaging with humans (see Table~\ref{tab:res-agent-chat}).
\end{itemize}

To enhance \qwen{}'s capabilities as an agent or copilot, we employ the self-instruct~\citep{self_instruct} strategy for SFT. Specifically, we utilize the in-context learning capability of \qwen{} for self-instruction. By providing a few examples, we can prompt \qwen{} to generate more relevant queries and generate outputs that follow a specific format, such as ReAct~\citep{yao2022react}. We then apply rules and involve human annotators to filter out any noisy samples. 
Afterwards, the samples are incorporated into \qwen{}'s training data, resulting in an updated version of \qwen{} that is more dependable for self-instruction.
We iterate through this process multiple times until we gather an ample number of samples that possess both exceptional quality and a wide range of diversity. As a result, our final collection consists of around \num{2000} high-quality samples. 

During the finetuning process, we mix these high-quality samples with all the other general-purpose SFT samples, rather than introducing an additional training stage. By doing so, we are able to retain essential general-purpose capabilities that are also pertinent for constructing agent applications.

\paragraph{Using Tools via ReAct Prompting}

We have created and made publicly available a benchmark for evaluating \qwen{}'s ability to call plugins, tools, functions, or APIs using ReAct Prompting~\citep[see][]{qwen-tool-eval}. 
To ensure fair evaluation, we have excluded any plugins that were included in \qwen{}'s training set from the evaluation set. The benchmark assesses the model's accuracy in selecting the correct plugin from a pool of up to five candidates, as well as the plausibility of the parameters passed into the plugin and the frequency of false positives. In this evaluation, a false positive occurs when the model incorrectly invokes a plugin in response to a query, despite not being required to do so.

The results presented in Table~\ref{tab:res-react-chat} demonstrate that \qwen{} consistently achieves higher accuracy in identifying the relevance of a query to the available tools as the model size increases. However, the table also highlights that beyond a certain point, there is little improvement in performance when it comes to selecting the appropriate tool and providing relevant arguments. This suggests that the current preliminary benchmark may be relatively easy and may require further enhancement in future iterations.
It is worth noting that GPT-3.5 stands out as an exception, displaying suboptimal performance on this particular benchmark. This could potentially be attributed to the fact that the benchmark primarily focuses on the Chinese language, which may not align well with GPT-3.5's capabilities. Additionally, we observe that GPT-3.5 tends to attempt to use at least one tool, even if the query cannot be effectively addressed by the provided tools.

\paragraph{Using Code Interpreter for Math Reasoning and Data Analysis}

The Python code interpreter is widely regarded as a powerful tool for augmenting the capabilities of an LLM agent. It is worth investigating whether \qwen{} can harness the full potential of this interpreter to enhance its performance in diverse domains, such as mathematical reasoning and data analysis. To facilitate this exploration, we have developed and made publicly available a benchmark that is specifically tailored for this purpose~\citep[see][]{qwen-code-interpreter-eval}.

The benchmark encompasses three primary categories of tasks: math problem-solving, data visualization, and other general-purpose tasks like file post-processing and web crawling. Within the visualization tasks, we differentiate between two levels of difficulty. The easier level can be achieved by simply writing and executing a single code snippet without the need for advanced planning skills. However, the more challenging level requires strategic planning and executing multiple code snippets in a sequential manner. This is because the subsequent code must be written based on the output of the previous code. For example, an agent may need to examine the structure of a CSV file using one code snippet before proceeding to write and execute additional code to create a plot.

Regarding evaluation metrics, we consider both the executability and correctness of the generated code. To elaborate on the correctness metrics, for math problems, we measure accuracy by verifying if the ground truth numerical answer is present in both the code execution result and the final response. When it comes to data visualization, we assess accuracy by utilizing \qwen{}\textsc{-VL}~\citep{qwenvl}, a powerful multimodal language model. \qwen{}\textsc{-VL} is capable of answering text questions paired with images, and we rely on it to confirm whether the image generated by the code fulfills the user's request.

The results regarding executability and correctness are presented in Table~\ref{tab:res-agent-code-executable-rate} and Table~\ref{tab:res-agent-code-correctness}, respectively. It is evident that \codellama{} generally outperforms \llamatwo{}, its generalist counterpart, which is not surprising since this benchmark specifically requires coding skills.
However, it is worth noting that specialist models that are optimized for code synthesis do not necessarily outperform generalist models. This is due to the fact that this benchmark encompasses various skills beyond coding, such as abstracting math problems into equations, understanding language-specified constraints, and responding in the specified format such as ReAct.
Notably, \qwensevenbchat{} and \qwenfourteenbchat{} surpass all other open-source alternatives of similar scale significantly, despite being generalist models.

\paragraph{Serving as a Hugging Face Agent}

Hugging Face provides a framework called the Hugging Face Agent or Transformers Agent~\citep{transformers-agents}, which empowers LLM agents with a curated set of multimodal tools, including speech recognition and image synthesis. This framework allows an LLM agent to interact with humans, interpret natural language commands, and employ the provided tools as needed.

To evaluate \qwen{}'s effectiveness as a Hugging Face agent, we utilized the evaluation benchmarks offered by Hugging Face. The results are presented in Table~\ref{tab:res-agent-chat}. The evaluation results reveal that \qwen{} performs quite well in comparison to other open-source alternatives, only slightly behind the proprietary GPT-4, demonstrating \qwen{}'s competitive capabilities.

\section{\qwencoder{}: Specialized Model for Coding}
\label{sec:coding}

Training on domain-specific data has been shown to be highly effective, particularly in the case of code pretraining and finetuning. A language model that has been reinforced with training on code data can serve as a valuable tool for coding, debugging, and interpretation, among other tasks. In this work, we have developed a series of generalist models using pretraining and alignment techniques. Building on this foundation, we have created domain-specific models for coding by leveraging the base language models of \qwen{}, including continued pretrained model, \qwencoder{} and supervised finetuned model, \qwencoderchat{}. Both models have \num{14} billion and \num{7} billion parameters versions.

\subsection{Code Pretraining}
We believe that relying solely on code data for pretraining can result in a significant loss of the ability to function as a versatile assistant. Unlike previous approaches that focused solely on pretraining on code data~\citep{AlphaCode, starcoder}, we take a different approach~\citep{code_llama} by starting with our base models \qwen{} trained on a combination of text and code data, and then continuing to pretrain on the code data. We continue to pretrain the models on a total of around \num{90} billion tokens.
During the pre-training phase, we initialize the model using the base language models \qwen{}.
Many applications that rely on specialized models for coding may encounter lengthy contextual scenarios, such as tool usage and code interpretation, as mentioned in Section \ref{sec:tool_use}. To address this issue, we train our models with context lengths of up to \num{8192}. 
Similar to base model training in Section \ref{sec:training}, we employ Flash Attention~\citep{flashattn} in the attention modules, and adopt the standard optimizer AdamW~\citep{adam,adamw}, setting  $\beta_{1} = 0.9$, $\beta_{2} = 0.95$, and $\epsilon = 10^{-8}$. We set the learning rate as $6.0 \times 10^{-5}$ for \qwencoderfourteenb~ and $3.0 \times 10^{-5}$ for \qwencodersevenb, with $3\%$ warm up iterations and no learning rate decays.

\subsection{Code Supervised Fine-Tuning}
After conducting a series of empirical experiments, we have determined that the multi-stage SFT strategy yields the best performance compared to other methods. 
In the supervised fine-tuning stage, the model \qwencoderchat{} initialized by the code foundation model \qwencoder{} are optimized by the AdamW~\citep{adam,adamw} optimizer ($\beta_{1} = 0.9$, $\beta_{2} = 0.95$, $\epsilon = 10^{-8}$) with a learning rate of $2.0 \times 10^{-6}$ and $1.0 \times 10^{-5}$ for the \num{14}B and \num{7}B model respectively. The learning rate increases to the peaking value with the cosine learning rate schedule ($3\%$ warm-up steps) and then remains constant.
\subsection{Evaluation}

\begin{table}[htbp]
    \centering
    \caption{Results of pass@1 (\%) on HumanEval and MBPP. Most scores are retrieved from the papers of StarCoder~\citep{starcoder}, CodeT5+~\citep{CodeT5+}, WizardCoder~\citep{wizardcoder} and \codellama{}~\citep{code_llama}. 
    }
    \begin{tabular}{lccc}
        \toprule
        \textbf{Model}  & \textbf{Params} & \textbf{HumanEval} & \textbf{MBPP} \\
        \arrayrulecolor{black}\midrule
        \multicolumn{4}{c}{\textit{Proprietary models}}\\
        \arrayrulecolor{black}\midrule
        PaLM & 540B & 26.2 & 36.8\\
        \arrayrulecolor{lightgray}\midrule
        PaLM-Coder & 540B & 36.0 & 47.0\\
        \arrayrulecolor{lightgray}\midrule
        PaLM 2-S & - & 37.6 & 50.0\\
        \arrayrulecolor{lightgray}\midrule
        Code-Cushman-001 & - & 33.5 & 45.9\\
        \arrayrulecolor{lightgray}\midrule
        Code-Davinci-002 & - & 47.0 & 58.1\\
        \arrayrulecolor{lightgray}\midrule
        GPT-3.5 & - & 73.2 & -\\
        \arrayrulecolor{lightgray}\midrule
        GPT-4 & - & 86.6 & -\\
        \arrayrulecolor{black}\midrule
        \multicolumn{4}{c}{\textit{Open-source models}}\\
        \arrayrulecolor{black}\midrule
        \multirow{4}{*}{\llamatwo{}} & 7B & 12.2 & 20.8\\
                     & 13B & 20.1 & 27.6\\
                     & \textcolor{gray}{34B} & \textcolor{gray}{22.6} & \textcolor{gray}{33.8} \\
                     & \textcolor{gray}{70B} & \textcolor{gray}{30.5} & \textcolor{gray}{45.4} \\
        \arrayrulecolor{lightgray}\midrule
        CodeGen-Multi & 16B & 18.3 & 20.9\\
        \arrayrulecolor{lightgray}\midrule
        CodeGen-Mono & 16B & 29.3 & 35.3\\
        \arrayrulecolor{lightgray}\midrule
        CodeGeeX2 & 6B & 35.9 & -\\
        \arrayrulecolor{lightgray}\midrule
        StarCoder-Prompted & 15B & 40.8 & 49.5\\
        \arrayrulecolor{lightgray}\midrule
        CodeT5+ & 16B & 30.9 & -\\
        \arrayrulecolor{lightgray}\midrule
        InstructCodeT5+ & 16B & 35.0 & -\\
        \arrayrulecolor{lightgray}\midrule
        \multirow{3}{*}{\codellama{}} & 7B & 33.5 & 41.4 \\
                     & 13B & 36.0 & 47.0 \\
                     & \textcolor{gray}{34B} & \textcolor{gray}{48.8} & \textcolor{gray}{55.0} \\
        \arrayrulecolor{lightgray}\midrule
        \multirow{3}{*}{\textsc{Code Llama-Instruct}} & 7B & 34.8 & 44.4 \\
                   & 13B & 42.7 & 49.4 \\ 
                   & \textcolor{gray}{34B} & \textcolor{gray}{41.5} & \textcolor{gray}{57.0} \\
        \arrayrulecolor{lightgray}\midrule
        \multirow{3}{*}{\textsc{Code Llama-Python}} &7B & 38.4 & 47.6\\
            & 13B & 43.3 & 49.0 \\
            & \textcolor{gray}{34B} & \textcolor{gray}{53.7} & \textcolor{gray}{56.2} \\
        \arrayrulecolor{lightgray}\midrule
        \textcolor{gray}{\textsc{Unnatural Code Llama}} & \textcolor{gray}{34B} & \textcolor{gray}{62.2} & \textcolor{gray}{61.2} \\
        \arrayrulecolor{lightgray}\midrule
        \multirow{2}{*}{WizardCoder-Python}	 & 13B & 64.0 & \bf 55.6 \\
                                 & \textcolor{gray}{34B} & \textcolor{gray}{73.2}	 & \textcolor{gray}{61.2} \\
         \arrayrulecolor{black}\midrule
        \multirow{2}{*}{\qwenchat{}}   & 7B & 37.2 & 35.8 \\ 
                          & 14B & 43.9 & 46.4 \\ 
        \arrayrulecolor{lightgray}\midrule
        \multirow{2}{*}{\qwencoder{}}     & 7B & 40.2  & 41.8 \\ 
                         & 14B & 45.1  & 51.4 \\ 
        \arrayrulecolor{lightgray}\midrule
        \multirow{2}{*}{\qwencoderchat{}} & 7B & 43.3  & 44.2 \\ 
                             & 14B  &  \bf 66.4     & 52.4 \\ 
        \arrayrulecolor{black}\bottomrule

    \end{tabular}            
    \label{tab:humaneval_mbpp}
\end{table}

\begin{table}[htbp]
\centering
\small
\caption{\textbf{Zero-shot pass@1 (\%) performance on the \textsc{HumanEvalPack} (synthesize) benchmark.} The baseline results are partly from \textsc{OctoPack}~\citep{octopack}.}
\label{tab:coding_humanevalpack_synthesize}
\resizebox{0.9\columnwidth}{!}{
\begin{tabular}{@{}lcccccccc@{}}
\toprule
\multirow{2}{*}[-\aboverulesep]{\textbf{Model}} & \multirow{2}{*}[-\aboverulesep]{\textbf{Params}} & \multicolumn{6}{c}{\textbf{Programming Language}}  & \\ \cmidrule(l){3-9}
             &   & Python & JavaScript   & Java    & Go      & C++  & Rust & Avg.  \\ 
\midrule
\multicolumn{9}{c}{\textit{Proprietary models}}\\
\midrule
GPT-4  & -  & 86.6 & 82.9& 81.7& 72.6& 78.7& 67.1   & 78.3 \\
\midrule
\multicolumn{9}{c}{\textit{Open-source models}}\\
\midrule
InstructCodeT5+   & 16B & 37.0& 18.9 &17.4& 9.5 &19.8& 0.3 &17.1  \\
\arrayrulecolor{lightgray}\midrule
StarChat-$\beta$  & 15B  & 33.5& 31.4 &26.7 &25.5 &26.6 &14.0& 26.3    \\
\arrayrulecolor{lightgray}\midrule
StarCoder         & 15B  & 33.6& 30.8 &30.2 &17.6 &31.6 &21.8 &27.6  \\
\arrayrulecolor{lightgray}\midrule
CodeGeeX2         & 6B  &35.9& 32.2& 30.8 &22.5 &29.3 &18.1 &28.1  \\
\arrayrulecolor{lightgray}\midrule
\textsc{OctoGeeX}         & 6B  & 44.7& 33.8& 36.9& 21.9 &32.3 &15.7& 30.9  \\
\arrayrulecolor{lightgray}\midrule
\textsc{OctoCoder}         &  15B & 46.2& 39.2& 38.2 &30.4 &35.6& 23.4 &35.5 \\ 
\arrayrulecolor{lightgray}\midrule
WizardCoder       &  15B & 59.8 &49.5& 36.1& 36.4& 40.9 &20.2 &40.5  \\ 
\arrayrulecolor{black}\midrule
\multirow{2}{*}{\qwenchat{}}      & 7B  & 37.2   &	23.2   & 32.9 & 20.7  & 22.0 & 9.1   &  24.2 \\
& 14B  &  43.9  &	38.4   & 42.7 & 34.1 & 24.4 & 18.9  &  33.7 \\
\arrayrulecolor{lightgray}\midrule
\multirow{2}{*}{\qwencoder{}}        & 7B  & 40.2   & 40.4   & 40.2 & 26.2 & 20.7 & 15.8   & 30.6  \\
         & 14B  & 45.1   & 51.8   & 57.3 & 39.6 &18.2 & 20.7   & 38.8  \\
\arrayrulecolor{lightgray}\midrule
\multirow{2}{*}{\qwencoderchat{}}     & 7B  & 43.3   & 41.5   & 49.4 & 29.3 & 32.9 & 20.1  & 36.1  \\
        & 14B  & \bf 66.4    & \bf 58.5   & \bf 56.1 & \bf 47.6 & \bf 54.2 & \bf 28.7   & \bf 51.9  \\
\arrayrulecolor{black}\bottomrule
\end{tabular}}
\label{tab:humanevalpack_synthesize}
\end{table}

Our \qwencoder{} models have been compared with both proprietary and open-source language models, as shown in Tables \ref{tab:humaneval_mbpp} and \ref{tab:humanevalpack_synthesize}. These tables present the results of our evaluation on the test sets of Humaneval~\citep{codex}, MBPP~\citep{mbpp}, and the multi-lingual code generation benchmark \textsc{HumanEvalPack}~\citep{octopack}. The comparison is based on the pass@1 performance of the models on these benchmark datasets. The results of this comparison are clearly demonstrated in Tables \ref{tab:humaneval_mbpp} and \ref{tab:humanevalpack_synthesize}.

Our analysis reveals that specialized models, specifically \qwencoder{} and \qwencoderchat{}, significantly outperform previous baselines with similar parameter counts, such as \textsc{OctoGeeX}~\citep{octopack}, InstructCodeT5+~\citep{CodeT5+}, and CodeGeeX2~\citep{CodeGeeX}. In fact, these models even rival the performance of larger models like Starcoder~\citep{starcoder}.

When compared to some of the extremely large-scale closed-source models, \qwencoder{} and \qwencoderchat{} demonstrate clear advantages in terms of pass@1. However, it is important to note that these models fall behind the state-of-the-art methods, such as GPT-4, in general. Nonetheless, with the continued scaling of both model size and data size, we believe that this gap can be narrowed in the near future.

It is crucial to emphasize that the evaluations mentioned previously are insufficient for grasping the full extent of the strengths and weaknesses of the models. In our opinion, it is necessary to develop more rigorous tests to enable us to accurately assess our relative performance in comparison to GPT-4.
\section{\mathqwen{}: Specialized Model for Mathematics Reasoning}

\label{sec:math}

\begin{table}[htbp]
\centering
\small
\caption{\textbf{Results of models on mathematical reasoning}. We report the accuracy of \qwen{} for all benchmarks using greedy decoding. For MATH, we are reporting \qwen{}'s performances on the test set from \cite{lightman2023lets}.}
\label{tab:res-math}
\begin{tabular}{@{}lccccc@{}}
\toprule
\textbf{Model} & \textbf{Params} & \textbf{GSM8K} & \textbf{MATH} & \textbf{Math401} & \textbf{Math23K} \\ 
\midrule
\multicolumn{6}{c}{\textit{Proprietary models}}\\
\arrayrulecolor{black}\midrule
GPT-4 & - &  \textbf{92.0} & \textbf{42.5} & 83.5 &  74.0 \\
\arrayrulecolor{lightgray}\midrule
GPT-3.5 & - & 80.8 & 34.1 & 75.1  & 60.0\\
\arrayrulecolor{lightgray}\midrule
\multirow{3}{*}{Minerva} & 8B & 16.2 & 14.1 & -& -    \\ 
 & 62B & 52.4 & 27.6 & - & -\\
 & 540B & 58.8 & 33.6 & -  & -\\
\arrayrulecolor{black}\midrule
\multicolumn{6}{c}{\textit{Open-source models}}\\
\arrayrulecolor{black}\midrule
\multirow{2}{*}{LLaMA-1 RFT} & 7B & 46.5 & 5.2 & -& -   \\
 & 13B & 52.1 & 5.1 & - & -   \\
\arrayrulecolor{lightgray}\midrule
\multirow{3}{*}{WizardMath} & 7B & 54.9 & 10.7  & - & -   \\
 & 13B & 63.9 & 14.0  & - & - \\
 & \textcolor{gray}{70B} & \textcolor{gray}{81.6} & \textcolor{gray}{22.7}   & \textcolor{gray}-& \textcolor{gray}-\\
 \arrayrulecolor{lightgray}\midrule
 \multirow{3}{*}{GAIRMath-Abel} & 7B & 59.7 & 13.0  & - & -   \\
 & 13B & 66.4 & 17.3   & -& -  \\
 & \textcolor{gray}{70B} & \textcolor{gray}{83.6} & \textcolor{gray}{28.3}   & \textcolor{gray}-& \textcolor{gray}-\\
\arrayrulecolor{black}\midrule
\multirow{2}{*}{\qwenchat{}} & 7B & 50.3 & 6.8 & 57.4  & 51.2 \\ 
 & 14B & 60.1 & 18.4 & 70.1 &67.0  \\ 
 \arrayrulecolor{lightgray}\midrule
\multirow{2}{*}{\mathqwenchat{}} & 7B & 62.5 & 17.2 & 80.8 & 75.4 \\ 
  & 14B & 69.8  & 24.2  & \textbf{85.0}  & \textbf{78.4}  \\ 
\arrayrulecolor{black}\bottomrule
\end{tabular}
\end{table}

We have created a mathematics-specialized model series called \mathqwenchat{}, which is built on top of the \qwen{} pretrained language models. Specifically, we have developed assistant models that are specifically designed to excel in arithmetic and mathematics and are aligned with human behavior.
We are releasing two versions of this model series, \mathqwenfourteenbchat{} and \mathqwensevenbchat{}, which have \num{14} billion and \num{7} billion parameters, respectively.

\subsection{Training}
We carry out math SFT on our augmented math instructional dataset for mathematics reasoning, and therefore we obtain the chat model, \mathqwenchat{}, directly. 
Owing to shorter average lengths of the math SFT data, we use a sequence length of \num{1024} for faster training. 
Most user inputs in the math SFT dataset are examination questions, and it is easy for the model to predict the input format and it is meaningless for the model to predict the input condition and numbers which could be random.
Thus, we mask the inputs of the system and user to avoid loss computation on them and find masking them accelerates the convergence during our preliminary experiments.
For optimization, we use the AdamW optimizer with the same hyperparameters of SFT except that we use a peak learning rate of $2 \times 10^{-5}$ and a training step of \num{50000}. 

\subsection{Evaluation}
We evaluate models on the test sets of GSM8K (Grade school math)~\citep{gsm8k}, MATH (Challenging competition math problems)~\citep{hendrycks2021measuring}, Math401 (Arithmetic ability)~\citep{math401}, and Math23K (Chinese grade school math)~\citep{Wang2017DeepNS}. We compare \mathqwenchat{} with proprietary models ChatGPT and Minerva~\citep{lewkowycz2022solving} and open-sourced math-specialized model RFT~\citep{rft}, WizardMath~\citep{wizardmath}, and GAIRMath-Abel~\citep{abel} in Table~\ref{tab:res-math}.
\mathqwenchat{} models show better math reasoning and arithmetic abilities compared to open-sourced models and \qwenchat{} models of similar sizes. 
Compared to proprietary models, \mathqwensevenbchat{} outperforms Minerva-8B in MATH.
\mathqwenfourteenbchat{} is chasing Minerva-62B and GPT-3.5 in GSM8K and MATH and delivers better performance on arithmetic ability and Chinese math problems.
\section{Related Work}
\label{sec:related-work}

\subsection{Large Language Models}
The excitement of LLM began with the introduction of the Transformer architecture~\citep{transformer}, which was then applied to pretraining large-scale data by researchers such as \citet{gpt, bert, roberta}. These efforts led to significant success in transfer learning, with model sizes growing from \num{100} million to over \num{10} billion parameters~\citep{t5, megatron}. 

In 2020, the release of GPT-3, a massive language model that is \num{10} times larger than T5, demonstrated the incredible potential of few-shot and zero-shot learning through prompt engineering and in-context learning, and later chain-of-thought prompting~\citep{cot}. This success has led to a number of studies exploring the possibilities of further scaling these models~\citep{bloom, opt,  glm, glm-130b, gshard, switch_transformer, glam, gpt-neo, gopher, chinchilla, palm, LaMDA}. As a result, the community has come to view these large language models as essential foundations for downstream models~\citep{foundation_models}. 

The birth of ChatGPT~\citep{chatgpt} and the subsequent launch of GPT-4~\citep{gpt4} marked two historic moments in the field of artificial intelligence, demonstrating that large language models (LLMs) can serve as effective AI assistants capable of communicating with humans. These events have sparked interests among researchers and developers in building language models that are aligned with human values and potentially even capable of achieving artificial general intelligence (AGI)~\citep{palm2, claude, claude2}.

One notable development in this area is the emergence of open-source LLMs, specifically LLaMA~\citep{llama} and \llamatwo{}~\citep{llama2}, which have been recognized as the most powerful open-source language models ever created. This has led to a surge of activity in the open-source community~\citep{huggingface}, with a series of large language models being developed collaboratively to build upon this progress~\citep{mpt, falcon, chatglm2, baichuan2, internlm}.

\subsection{Alignment}

The community was impressed by the surprising effectiveness of alignment on LLMs. Previously, LLMs without alignment often struggle with issues such as repetitive generation, hallucination, and deviation from human preferences. Since 2021, researchers have been diligently working on developing methods to enhance the performance of LLMs in downstream tasks~\citep{flan, t0, flan-collection, flanv2, bloomz}. Furthermore, researchers have been actively exploring ways to align LLMs with human instructions~\citep{instructgpt, askell2021general, PMP, cai}. 
One major challenge in alignment research is the difficulty of collecting data. While OpenAI has utilized its platform to gather human prompts or instructions, it is not feasible for others to collect such data. 

However, there has been some progress in this area, such as the self-instruct approach proposed in~\cite {self_instruct}. This innovative work offers a potential solution to the data collection problem in alignment research. As a result, there has been a surge in open-source chat data, including Alpaca~\citep{alpaca}, MOSS~\citep{moss}, Dolly~\citep{dolly}, Evol-Instruct~\citep{wizardlm}, and others~\citep{self_align, expertprompting, baize, phoenix, ultrachat, belle, firefly}. Similarly, there has been an increase in open-source chat models, such as Alpaca~\citep{alpaca}, Vicuna~\citep{vicuna}, Guanaco~\citep{qlora}, MOSS~\citep{moss},  WizardLM~\citep{wizardlm}, and others~\citep{baize, phoenix, ultrachat, tulu}.

To train an effective chat model, available solutions are mostly based on SFT and RLHF~\citep{instructgpt}. 
While SFT is similar to pretraining, it focuses on instruction following using the aforementioned data. However, for many developers, the limited memory capacity is a major obstacle to further research in SFT. As a result, parameter-efficient tuning methods, such as LoRA~\citep{lora} and Q-LoRA~\citep{qlora}, have gained popularity in the community. LoRA tunes only low-rank adapters, while Q-LoRA builds on LoRA and utilizes 4-bit quantized LLMs and paged attention~\citep{bnb, gptq, pageattention}. In terms of RLHF, recent methods such as PPO~\citep{ppo, llama2} have been adopted, but there are also alternative techniques aimed at addressing the complexity of optimization, such as RRHF~\citep{yuan2023rrhf}, DPO~\citep{dpo}, and PRO~\citep{pro}. Despite the ongoing debate about the effectiveness of RLHF, more evidence is needed to understand how it enhances the intelligence of LLMs and what potential drawbacks it may have.

\subsection{Tool Use and Agents} 

LLM's planning function allows for the invocation of tools, such as APIs or agent capabilities, through in-context learning, as demonstrated by \citet{toolformer}. \citet{yao2022react} introduced ReAct, a generation format that enables the model to generate thoughts on which tool to use, accept input from API observations, and generate a response. GPT-3.5 and GPT-4, when prompted with few shots, have shown consistent and impressive performance.
In addition to tool usage, LLMs can utilize external memory sources like knowledge bases~\citep{chatdb, memorybank} or search engines~\citep{webgpt, webglm} to generate more accurate and informative answers. This has led to the popularity of frameworks like LangChain~\citep{langchain}.
The research on LLMs for tool use has also sparked interest in building agents with LLM capabilities, such as agents that can call different AI models~\citep{hugginggpt, modelscope_agent}, embodied lifelong learning or multimodal agents~\citep{voyager, palm-e}, and multiple agents interacting with each other and even building a micro-society~\citep{agentverse, camel, werewolf, metagpt}.

\subsection{LLM for Coding}

Previous research has demonstrated that LLMs possess remarkable capabilities in code understanding and generation, particularly those with massive numbers of parameters~\citep{palm, palm2, gopher, chinchilla}. 
Moreover, several LLMs have been pre-trained, continued pre-trained, or fine-tuned on coding-related data, which has resulted in significantly improved performance compared to general-purpose LLMs. These models include Codex~\cite{codex}, AlphaCode~\citep{AlphaCode}, SantaCoder~\citep{santacoder}, Starcoder-Base~\citep{starcoder}, InCoder~\citep{incoder}, CodeT5~\citep{codet5}, CodeGeeX~\citep{CodeGeeX}, and \codellama{}~\citep{code_llama}.
In addition to these models, recent studies have focused on developing specialized alignment techniques for coding, such as Code Llama-Instruct~\citep{code_llama} and StarCoder~\citep{starcoder}. These models can assist developers in various code-related tasks, including code generation~\citep{codex,mbpp}, code completion~\citep{repocoder}, code translation~\citep{code_translation_compiler}, bug fixing~\citep{octopack}, code refinement~\citep{code_refinement}, and code question answering~\citep{code_qa}.
In a word, LLMs have the potential to revolutionize the field of coding by providing developers with powerful tools for code comprehension, generation, and related tasks.

\subsection{LLM for Mathematics}

LLMs with a certain model scale have been found to possess the ability to perform mathematical reasoning~\citep{Wei2022EmergentAO, bbh}. In order to encourage LLMs to achieve better performance on math-related tasks, researchers have employed techniques such as chain-of-thought prompting~\citep{cot} and scratchpad~\citep{scratchpad}, which have shown promising results. Additionally, self-consistency~\citep{self_consistency} and least-to-most prompting~\citep{least_to_most} have further improved the performance of these models on these tasks.
However, prompt engineering is a time-consuming process that requires a lot of trial and error, and it is still difficult for LLMs to consistently perform well or achieve satisfactory results in solving mathematical problems. Moreover, simply scaling the data and model size is not an efficient way to improve a model's mathematical reasoning abilities. 
Instead, pretraining on math-related corpora has been shown to consistently enhance these capabilities~\citep{hendrycks2021measuring,lewkowycz2022solving,taylor2022galactica,lightman2023lets}. Additionally, fine-tuning on math-related instruction-following datasets~\citep{alpaca-cot, rft, wizardmath, yue2023mammoth,abel,yu2023metamath}, has also been effective and more cost-effective than math-specific pretraining. 
Despite their limitations in terms of accuracy, LLMs still have significant potential to assist users with practical mathematical problems. There is ample scope for further development in this area.
\section{Conclusion}
\label{sec:conclusion}

In this report, we present the \qwen{} series of large language models, which showcase the latest advancements in natural language processing. With 14B, 7B, and 1.8B parameters, these models have been pre-trained on massive amounts of data, including trillions of tokens, and fine-tuned using cutting-edge techniques such as SFT and RLHF. Additionally, the \qwen{} series includes specialized models for coding and mathematics, such as \qwencoder{}, \qwencoderchat{}, and \mathqwenchat{}, which have been trained on domain-specific data to excel in their respective fields.
Our results demonstrate that the \qwen{} series is competitive with existing open-source models and even matches the performance of some proprietary models on comprehensive benchmarks and human evaluation. 

We believe that the open access of \qwen{} will foster collaboration and innovation within the community, enabling researchers and developers to build upon our work and push the boundaries of what is possible with language models. By providing these models to the public, we hope to inspire new research and applications that will further advance the field and contribute to our understanding of the variables and techniques introduced in realistic settings.
In a nutshell, the \qwen{} series represents a major milestone in our development of large language models, and we are excited to see how it will be used to drive progress and innovation in the years to come.

\clearpage

\bibliography{biblio}
\bibliographystyle{iclr2023_conference}
\clearpage

\appendix
\section{Appendix}

\subsection{More Training Details}

\subsubsection{Data Format for \qwenchat{}}
\label{appendix:data_format}
Different from conventional pretraining based on autoregressive next-token prediction, despite using a similar training task, there should be a specially design data format for SFT and RLHF to build a conversational AI assistant model. 
Common formats include ``human-assistant'' and ChatML formats. 
As to our knowledge, one of the earliest examples of the human-assistant format comes from Anthropic~\citep{PMP}, which adds a special phrase ``\verb|\n\nhuman: |'' in front of the user input and ``\verb|\n\nassistant: |'' in front of the assistant response. It is easy for the base language model to transfer to the pattern of conversational AI. However, as the specific phrases are common words, it might be hard for the model to disambiguate from these words in other contexts. 

Instead, we turned to the ChatML format proposed by OpenAI.\footnote{https://github.com/openai/openai-python/blob/main/chatml.md} This format allows the use of special tokens, i.e., ``\verb|<im_start>|'' and ``\verb|<im_end>|'', that do not appear in pretraining, and thus resolve the aforementioned problem. We demonstrate an example of the format below.

\tcbset{
    breakable, enhanced,   
    left*=10pt, right*=10pt,
    top=0pt, bottom=0pt,
    colback=white!10!white,
    colframe=black!75!black,
    fonttitle=\bfseries\large,
    subtitle style={boxrule=0pt,colback=gray!50!white},
}
\begin{tcolorbox}[title=ChatML Format]
\small
\begin{lstlisting}
<|im_start|>system
You are a helpful assistant.<|im_end|>
<|im_start|>user
Hello!<|im_end|>
<|im_start|>assistant
Hello! How can I assist you today?<|im_end|>
\end{lstlisting}
\end{tcolorbox}

\subsection{Evaluation}

\subsubsection{Automatic Evaluation}

To provide a whole picture of the performance of our model series \qwen{}, here in this section we illustrate the detailed performance of our models as well as the baselines in the comprehensive benchmark evaluation proposed by \citet{opencompass}.
We report the results in multiple tables based on the officially provided categories, including examination, language, knowledge, understanding, and reasoning. In terms of the performance of the baseline models, we report the higher results between the reported ones and those on the leaderboard. 

\paragraph{Examination} Here we evaluate the models on a series of datasets relevant to the examination. The datasets include:

\begin{itemize}
    \item \textbf{MMLU} \citep{mmlu} Massive Multi-task Language Understanding is designed for measuring language understanding capabilities. We report 5-shot results.
    \item \textbf{C-Eval} \citep{ceval} C-Eval is a Chinese evaluation dataset spanning \num{52} diverse disciplines. We report 5-shot results.
    \item \textbf{CMMLU} \citep{cmmlu} CMMLU is designed for assessing language understanding capabilities in Chinese.  We report 5-shot results.
    \item \textbf{AGIEval} \citep{agieval} This is a benchmark consisting of human-centric examinations, including college entrance exams, law school admission tests, math competitions, and lawyer qualification tests. We report zero-shot results.
    \item \textbf{Gaokao-Bench} \citep{gaokao-bench} This is a benchmark with Gaokao (Chinese college-entrance examination) questions.  We report zero-shot results.
    \item \textbf{ARC} \citep{arc} ARC is a dataset consisting of grade-school level, multiple-choice science questions. It includes an easy set and a challenge set, which are referred by ARC-e and ARC-c. We report zero-shot results.
\end{itemize}

\begin{table}[htbp]
\centering
\small
\caption{\textbf{Results on MMLU}. All are tested with five-shot accuracy. We provide the reported results of the other models for comparison.}
\label{tab:exam-mmlu}
\begin{tabular}{@{}lcccccc@{}}
\toprule
\bf Model & \bf Params &
  \multicolumn{1}{c}{\bf Average} &
  \multicolumn{1}{c}{\bf STEM} &
  \multicolumn{1}{c}{\bf Social Sciences} &
  \multicolumn{1}{c}{\bf Humanities} &
  \multicolumn{1}{c}{\bf Others} \\
  \midrule
\multirow{2}{*}{MPT}  & 7B & 26.8     & 25.3 & 27.1 & 26.7 & 28.2 \\
 & \textcolor{gray}{30B} & \textcolor{gray}{46.9}     & \textcolor{gray}{39.0} & \textcolor{gray}{52.8} & \textcolor{gray}{44.5} & \textcolor{gray}{52.9} \\
 \arrayrulecolor{lightgray}\midrule
 \multirow{2}{*}{Falcon}  & 7B & 26.2 & 26.2 & 24.7 & 26.4 & 27.4 \\
 & \textcolor{gray}{40B} & \textcolor{gray}{55.4}     & \textcolor{gray}{45.5} & \textcolor{gray}{65.4} & \textcolor{gray}{49.3} & \textcolor{gray}{65.0} \\
 \arrayrulecolor{lightgray}\midrule
\multirow{2}{*}{ChatGLM2}  & 6B & 47.9     & 41.2 & 54.4 & 43.7 & 54.5 \\
 & 12B & 56.2     & 48.2 & 65.1 & 52.6 & 60.9 \\
 \arrayrulecolor{lightgray}\midrule
InternLM  & 7B & 51.0     & -    & -    & -    & -    \\
\arrayrulecolor{lightgray}\midrule
\multirow{2}{*}{Baichuan2} & 7B & 54.2     & -    & -    & -    & -    \\
 & 13B & 59.2     & -    & -    & -    & -    \\
 \arrayrulecolor{lightgray}\midrule
XVERSE   & 13B & 55.1     & 44.5 & 64.4 & 50.5 & 62.9 \\
\arrayrulecolor{lightgray}\midrule
\multirow{4}{*}{LLaMA}    & 7B & 35.1     & 30.5 & 38.3 & 34.0 & 38.1 \\
  & 13B   & 46.9     & 35.8 & 53.8 & 45.0 & 53.3 \\
  & \textcolor{gray}{33B}   & \textcolor{gray}{57.8}     & \textcolor{gray}{46.0} & \textcolor{gray}{66.7} & \textcolor{gray}{55.8} & \textcolor{gray}{63.4} \\
  & \textcolor{gray}{65B}   & \textcolor{gray}{63.4}     & \textcolor{gray}{51.7} & \textcolor{gray}{72.9} & \textcolor{gray}{61.8} & \textcolor{gray}{67.4} \\
\arrayrulecolor{lightgray}\midrule
\multirow{4}{*}{\llamatwo{}}   & 7B  & 45.3     & 36.4 & 51.2 & 42.9 & 52.2 \\
   & 13B & 54.8     & 44.1 & 62.6 & 52.8 & 61.1 \\
   & \textcolor{gray}{34B}   & \textcolor{gray}{62.6}     & \textcolor{gray}{52.1} & \textcolor{gray}{71.8} & \textcolor{gray}{59.4} & \textcolor{gray}{69.2} \\
   & \textcolor{gray}{70B}   & \textcolor{gray}{68.9}     & \textcolor{gray}{58.0} & \textcolor{gray}{80.3} & \textcolor{gray}{65.0} & \textcolor{gray}{74.6} \\
   \arrayrulecolor{black}\midrule
\multirow{3}{*}{\qwen{}} & 1.8B & 44.6     & 39.6 & 50.0 & 40.4 & 51.0 \\ 
  & 7B & 58.2     & 50.2 & 68.6 & 52.5 & 64.9 \\
 & 14B & \bf 66.3     & \bf 59.4 & \bf 76.2 & \bf 60.9 & \bf 71.8 \\ 
 \arrayrulecolor{black}\bottomrule
\end{tabular}
\end{table}

\begin{table}[tb]
\centering
\small
\caption{\textbf{Leaderboard results of C-Eval}. We include the results of both proprietary models and open-source models. Note that there are a number of models on the leaderboard with very few details, in terms of proprietary models, we only report the results of GPT-3.5, GPT-4, InternLM and ChatGLM2.
}
\label{tab:exam-ceval}
\begin{tabular}{@{}lccccccc@{}}
\toprule
Model & Params & Avg. & Avg. (Hard) & STEM & Social Sciences & Humanities & Others \\ 
\midrule
\multicolumn{8}{c}{\textit{Proprietary models}}\\
\midrule
GPT-3.5 & - & 54.4 & 41.4 & 52.9 & 61.8 & 50.9 & 53.6 \\
\arrayrulecolor{lightgray}\midrule
GPT-4 & - & 68.7 & \textbf{54.9} & \textbf{67.1} & 77.6 & 64.5 & 67.8 \\
\arrayrulecolor{lightgray}\midrule
InternLM & 123B & 68.8 & 50.0 & 63.5 & 81.4 & 72.7 & 63.0 \\
\arrayrulecolor{lightgray}\midrule
ChatGLM2 & - & \textbf{71.1} & 50.0 & 64.4 & \textbf{81.6} & \textbf{73.7} & \textbf{71.3} \\
\arrayrulecolor{black}\midrule
\multicolumn{8}{c}{\textit{Open-source models}}\\
\arrayrulecolor{black}\midrule
ChatGLM2 & 6B             & 51.7     & 37.1 & 48.6 & 60.5 & 51.3 & 49.8 \\
\arrayrulecolor{lightgray}\midrule
InternLM & 7B             & 52.8     & 37.1 & 48.0 & 67.4 & 55.4 & 45.8 \\
\arrayrulecolor{lightgray}\midrule
\multirow{2}{*}{Baichuan2}  & 7B    & 54.0     & - & - & - & - & - \\
        & 13B   & 58.1     & - & - & - & - & - \\
\arrayrulecolor{lightgray}\midrule
XVERSE     & 13B         & 54.7     & 33.5 & 45.6 & 66.2 & 58.3 & 56.9 \\
\arrayrulecolor{black}\midrule
\multirow{3}{*}{\qwen{}}   & 1.8B        & 54.7     & 41.8 & 50.8 & 69.9 & 56.3 & 46.2 \\
 & 7B         & 63.5     & 46.4 & 57.7 & 78.1 & 66.6 & 57.8 \\
 & 14B        & \bf 72.1     & \bf 53.7 & \bf 65.7 & \bf 85.4 & \bf 75.3 & \bf 68.4 \\ \bottomrule
\end{tabular}
\end{table}

\begin{table}[tbp]
\centering
\small
\caption{\textbf{Results on the other datasets of examination}. Specifically, we report the results on CMMLU, AGIEval, ARC-e, and ARC-c.}
\label{tab:exam-others}
\begin{tabular}{@{}lcccccc@{}}
\toprule
\bf Model & \bf Params &
  \multicolumn{1}{c}{\bf CMMLU} &
  \multicolumn{1}{c}{\bf AGIEval} &
  \multicolumn{1}{c}{\bf Gaokao-Bench} &
  \multicolumn{1}{c}{\bf ARC-e} &
  \multicolumn{1}{c}{\bf ARC-c} \\
  \midrule
\multirow{1}{*}{MPT}  & 7B & 25.9 & 21.3 & 19.8 & 70.2 & 42.6 \\
 \arrayrulecolor{lightgray}\midrule
 \multirow{1}{*}{Falcon}  & 7B & - & - & - & 70.0 & 42.4 \\
  \arrayrulecolor{lightgray}\midrule
\multirow{1}{*}{ChatGLM2}  & 6B & 49.3   & 39.0 & 46.4 & 73.0 & 61.0 \\
 \arrayrulecolor{lightgray}\midrule
\multirow{2}{*}{InternLM}  & 7B & 51.8     & 36.9    & 43.0    & 78.7    & 69.5    \\
& \textcolor{gray}{20B} & \textcolor{gray}{59.0} & \textcolor{gray}{44.6} & \textcolor{gray}{45.5} & \textcolor{gray}{86.1} & \textcolor{gray}{81.7} \\
\arrayrulecolor{lightgray}\midrule
\multirow{2}{*}{Baichuan2} & 7B & 57.1    & 42.7    & 47.5    & 54.7    & 32.5    \\
 & 13B & 62.0     & 48.2    & 54.3    & 61.9    & 38.0    \\
 \arrayrulecolor{lightgray}\midrule
\multirow{4}{*}{LLaMA}    & 7B & 26.8     & 20.6 & 21.3 & 72.8 & 47.6 \\
  & 13B   & 31.5     & 22.0 & 20.4 & 74.8 & 52.7 \\
  & \textcolor{gray}{33B}   & \textcolor{gray}{36.0}     & \textcolor{gray}{33.5} & \textcolor{gray}{18.9} & \textcolor{gray}{80.0} & \textcolor{gray}{67.5} \\
  & \textcolor{gray}{65B}   & \textcolor{gray}{40.6}     & \textcolor{gray}{33.9} & \textcolor{gray}{19.1} & \textcolor{gray}{80.6} & \textcolor{gray}{69.5} \\
\arrayrulecolor{lightgray}\midrule
\multirow{3}{*}{\llamatwo{}}   & 7B  & 31.8     & 21.8 & 18.9 & 75.2 & 45.9 \\
   & 13B & 38.4     & 30.9 & 18.2 & 77.3 & 60.3 \\
   & \textcolor{gray}{70B}   & \textcolor{gray}{53.6}     & \textcolor{gray}{40.2} & \textcolor{gray}{23.3} & \textcolor{gray}{85.9} & \textcolor{gray}{78.3} \\
\arrayrulecolor{lightgray}\midrule
StableBeluga2 &  \textcolor{gray}{70B} & \textcolor{gray}{51.8} & \textcolor{gray}{41.6} & \textcolor{gray}{40.9} & \textcolor{gray}{91.2} & \textcolor{gray}{86.1} \\
   \arrayrulecolor{black}\midrule
\multirow{3}{*}{\qwen{}} & 1.8B & 49.3     & 36.9 & 44.9 & 71.6 & 53.2 \\ 
  & 7B & 62.2     & 45.8 & 52.5 & 84.0 & 75.3 \\
 & 14B & \bf 71.0     & \bf 52.3 & \bf 61.9 & \bf 90.3 & \bf 84.4\\ 
 \arrayrulecolor{black}\bottomrule
\end{tabular}
\end{table}

In terms of MMLU, we report the detailed results in Table~\ref{tab:exam-mmlu}. In terms of C-Eval, we report the results in Table~\ref{tab:exam-ceval}. For the rest of the datasets, we report the results in Table~\ref{tab:exam-others}. Note that AGIEval includes the parts of Chinese and English, while \llamatwo{} only reported the results in the English part, so we use the results on OpenCompass. Additionally, while CMMLU, AGIEval, and Gaokao-Bench are related to Chinese, and MPT, Falcon, and the LLaMA series were not optimized for Chinese, these models achieved low performance on the datasets. 

\paragraph{Knowledge and Understanding} Here we evaluate the models on a series of datasets relevant to knowledge and natural language understanding.
The datasets include

\begin{itemize}
    \item \textbf{BoolQ} \citep{boolq} This is a QA dataset, where the questions are about passages of Wikipedia, and the model should answer yes or no to the given possible answer. We report zero-shot results.
    \item \textbf{CommonsenseQA} \citep{commonsenseqa} This is a dataset of multiple-choice question answering that asseses the understanding of commonsense knowledge. We report 8-shot results.
    \item \textbf{NaturalQuestions} \citep{naturalquestions} It is a dataset of QA where the questions are from users and the answers are verified by experts. We report zero-shot results.
    \item \textbf{LAMBADA} \citep{lambada} This is dataset to evaluate language understanding by word prediction. It consists of passages related to human subjects. We report zero-shot results.
\end{itemize}

\begin{table}[tbp]
\centering
\small
\caption{\textbf{Results on the datasets concerning knowledge and understanding.} Specifically, we report the results on BoolQ, CommonsenseQA, NaturalQuestions, and LAMBADA.}
\label{tab:knowledge}
\begin{tabular}{@{}lccccc@{}}
\toprule
\bf Model & \bf Params &
  \multicolumn{1}{c}{\bf BoolQ} &
  \multicolumn{1}{c}{\bf CommonsenseQA} & 
  \multicolumn{1}{c}{\bf NaturalQuestions} &
  \multicolumn{1}{c}{\bf LAMBADA} \\
  \midrule
\multirow{2}{*}{MPT}  & 7B & 75.0 & 61.8 & 11.6 & 70.0 \\
 \arrayrulecolor{lightgray}\midrule
 \multirow{2}{*}{Falcon}  & 7B & 67.5 & 20.8 & 15.7 & -\\
 \arrayrulecolor{lightgray}\midrule
\multirow{1}{*}{ChatGLM2}  & 6B & 79.0   & 65.4 & 9.7 & 54.3\\
 \arrayrulecolor{lightgray}\midrule
\multirow{2}{*}{InternLM}  & 7B & 64.1     & 59.8       & 8.9     & 67.0 \\
& \textcolor{gray}{20B} & \textcolor{gray}{87.5} & \textcolor{gray}{70.6} & \textcolor{gray}{25.2} & \textcolor{gray}{71.8}  \\
\arrayrulecolor{lightgray}\midrule
XVERSE   & 13B & 64.2     & 62.2 & 0.3  & 48.2 \\
\arrayrulecolor{lightgray}\midrule
\multirow{2}{*}{Baichuan2} & 7B & 63.2    & 63.0     & 9.4    & 73.3  \\
 & 13B & 67.0     & 65.6       & 16.3     & 74.0 \\
 \arrayrulecolor{lightgray}\midrule
\multirow{4}{*}{LLaMA}    & 7B & 76.5     & 64.9  & 16.8 & 73.3 \\
  & 13B   & 78.7     & 67.4  & 20.2 & 75.2 \\
  & \textcolor{gray}{33B}   & \textcolor{gray}{84.4}     & \textcolor{gray}{72.5}  & \textcolor{gray}{30.9} & \textcolor{gray}{77.2}  \\
  & \textcolor{gray}{65B}   & \textcolor{gray}{86.6}     & \textcolor{gray}{74.1}  & \textcolor{gray}{33.4} & \textcolor{gray}{77.7} \\
\arrayrulecolor{lightgray}\midrule
\multirow{3}{*}{\llamatwo{}}   & 7B  & 77.4     & 66.5  & 19.1 & 73.3 \\
   & 13B & 82.4    & 67.3  & \bf 24.9 & \bf 76.5 \\
   & \textcolor{gray}{70B}   & \textcolor{gray}{87.7}     & \textcolor{gray}{78.5}  & \textcolor{gray}{34.2} & \textcolor{gray}{78.9}  \\
\arrayrulecolor{lightgray}\midrule
StableBeluga2 &  \textcolor{gray}{70B} & \textcolor{gray}{89.4} & \textcolor{gray}{72.6} & \textcolor{gray}{25.1} & \textcolor{gray}{71.3}  \\
   \arrayrulecolor{black}\midrule
\multirow{3}{*}{\qwen{}} & 1.8B & 68.0     & 60.1  & 3.2 & 58.4\\ 
  & 7B & 76.4     & 66.8  & 17.4  & 67.9 \\
 & 14B & \bf 86.2     & \bf 70.3 &  23.9 & 71.1 \\ 
 \arrayrulecolor{black}\bottomrule
\end{tabular}
\end{table}

We report the results in Table~\ref{tab:knowledge}.

\paragraph{Reasoning} We report the evaluation results on the datasets concerning reasoning, focusing on natural language reasoning. For the others, such as mathematics and coding, as we have illustrated detailed results, here we do not report those results repeatedly. The datasets for evaluation include:

\begin{itemize}
    \item \textbf{HellaSwag} \citep{hellaswag} This is a commonsense natural language inference (NLI) dataset, where the questions are easy for humans but struggling for previous language models. We report zero-shot results.
    \item \textbf{PIQA} \citep{piqa} This is an NLI dataset assessing the physical knowledge. We report zero-shot results.
    \item \textbf{SIQA} \citep{siqa} This is an NLI dataset evaluating social commonsense intelligence. We report zero-shot results.
    \item \textbf{OCNLI} \citep{ocnli} This is an NLI dataset focusing on Chinese. We report zero-shot results.
\end{itemize}

\begin{table}[tbp]
\centering
\small
\caption{\textbf{Results on the datasets related to natural language reasoning.} Specifically, we report the results on HellaSwag, PIQA, SIQA, and OCNLI.}
\label{tab:reasoning}
\begin{tabular}{@{}lccccc@{}}
\toprule
\bf Model & \bf Params &
  \multicolumn{1}{c}{\bf HellaSwag} &
  \multicolumn{1}{c}{\bf PIQA} & 
  \multicolumn{1}{c}{\bf SIQA} &
  \multicolumn{1}{c}{\bf OCNLI} \\
  \midrule
\multirow{1}{*}{MPT}  & 7B & 76.4 & \bf 80.6 & 48.5 & 30.0   \\
 \arrayrulecolor{lightgray}\midrule
 \multirow{1}{*}{Falcon}  & 7B & 74.1 & 76.7 & 47.2 & -  \\
 \arrayrulecolor{lightgray}\midrule
\multirow{1}{*}{ChatGLM2}  & 6B & 57.0   & 69.6 & 64.3  & 33.1  \\ \arrayrulecolor{lightgray}\midrule
\multirow{2}{*}{InternLM}  & 7B & 70.6     & 77.9       & 60.5   & 37.5     \\
& \textcolor{gray}{20B} & \textcolor{gray}{78.1} & \textcolor{gray}{80.3} & \textcolor{gray}{72.8} & \textcolor{gray}{42.5}  \\
\arrayrulecolor{lightgray}\midrule
\multirow{2}{*}{Baichuan2} & 7B & 67.0    & 76.2     & 44.4 & 30.3       \\
 & 13B & 70.8     & 78.1       & 44.3    & 30.0   \\
 \arrayrulecolor{lightgray}\midrule
\multirow{4}{*}{LLaMA}    & 7B & 76.1     & 79.8  & 48.9  & 33.6  \\
  & 13B   & 79.2     & 80.1  & 52.5  & 32.1 \\
  & \textcolor{gray}{33B}   & \textcolor{gray}{82.8}     & \textcolor{gray}{82.3}  & \textcolor{gray}{57.8} & \textcolor{gray}{30.7} \\
  & \textcolor{gray}{65B}   & \textcolor{gray}{84.2}     & \textcolor{gray}{82.8}  & \textcolor{gray}{61.2} & \textcolor{gray}{44.9} \\
\arrayrulecolor{lightgray}\midrule
\multirow{3}{*}{\llamatwo{}}   & 7B  & 77.2     & 78.8  & 48.5   & 32.1 \\
   & 13B & \bf 80.7    & 80.5  &  54.8   & 34.1   \\
   & \textcolor{gray}{70B}   & \textcolor{gray}{85.3}     & \textcolor{gray}{82.8}  & \textcolor{gray}{64.8} & \textcolor{gray}{46.5}  \\
\arrayrulecolor{lightgray}\midrule
StableBeluga2 & \textcolor{gray}{70B} & \textcolor{gray}{84.1} & \textcolor{gray}{83.3} & \textcolor{gray}{78.1} & \textcolor{gray}{48.3} \\
   \arrayrulecolor{black}\midrule
\multirow{3}{*}{\qwen{}} & 1.8B & 56.7     & 73.3  & 56.1 & 39.0 \\ 
  & 7B & 75.1     & 77.9  & 69.9  & 47.4  \\
 & 14B & 80.2    & 79.9 &  \bf 77.9  & \bf 57.9 \\ 
 \arrayrulecolor{black}\bottomrule
\end{tabular}
\end{table}

We report the results in Table~\ref{tab:reasoning}.

\subsubsection{Human Evaluation}\label{app:cases}
In this section, we demonstrate the cases of human analysis. 
In our self-constructed evaluation dataset, the instructions are either manually written data or manual revised from public datasets, such as CLiB\footnote{https://github.com/jeinlee1991/chinese-llm-benchmark}, C-Eval~\citep{ceval}, FacTool~\citep{chern2023factool}, LeetCode\footnote{https://leetcode.cn/}), etc.

In terms of each case, we demonstrate the responses and Elo ratings\footnote{https://en.wikipedia.org/wiki/Elo\_rating\_system} of all models for comparison. Specifically, as the data in our human evaluation are in Chinese, we also provide their translations in English.

\tcbset{
    breakable, enhanced,   
    left*=10pt, right*=10pt,
    top=0pt, bottom=0pt,
    colback=white!10!white,
    colframe=black!75!black,
    fonttitle=\bfseries\large,
    subtitle style={boxrule=0pt,colback=gray!50!white},
}
\lstset{language=python, breaklines=true}
\begin{CJK*}{UTF8}{gbsn}
\begin{tcolorbox}[title=Knowledge]
\small
北京大学的前身是什么学校？\\
What was the predecessor of Peking University?
\tcbsubtitle{Responses}
\setlength{\tabcolsep}{0.1mm}{
\begin{xltabular}{\linewidth}{
    p{0.25\textwidth}<{\raggedright}
    X
    p{0.1\textwidth}<{\centering}
    }
    \textbf{Model} & \textbf{Response} & \textbf{Elo} \\
\textbf{Qwen-7B-Chat (SFT)}  & 北京大学的前身是京师大学堂，它创建于1898年，是中国近代第一所国立大学，也是最早以“大学”
命名的高等学府。& 1037 \\
& The predecessor of Peking University was the Imperial University of Peking, established in 1898. It was the first national university in modern China and also the earliest higher education institution to be named a ``university''. & \\
\\[3pt]
\textbf{Qwen-14B-Chat (SFT)}  & 北京大学的前身是京师大学堂，创建于1898年，是中国近代第一所国立大学。 & 1011 \\
& The predecessor of Peking University is the Imperial University of Peking, founded in 1898 and the first national university in modern China. & \\
\\[3pt]
\textbf{Qwen-14B-Chat (RLHF)}  & 北京大学的前身是京师大学堂，创建于1898年，是中国近代第一所国立综合性大学。京师大学堂最初设立的学科有经学、文学、史学、哲学、法律、教育、农学、工学、商学、医学等，后来逐渐扩大到社会科学、自然科学、工程技术、艺术等多个领域。1912年，京师大学堂改名为北京大学，成为中国最高学府之一，也是中国现代高等教育的发源地之一。 & 1090 \\
& Peking University, originally known as the Imperial University of Peking, was founded in 1898 and is the first comprehensive national university in modern China. The initial disciplines established at the Imperial University of Peking included classics, literature, history, philosophy, law, education, agriculture, engineering, business, medicine, etc., gradually expanding to encompass social sciences, natural sciences, engineering and technology, arts, and other fields. In 1912, the Imperial University of Peking changed its name to Peking University, becoming one of the top universities in China and also one of the birthplaces of modern higher education in China. & \\
\\[30pt]
\textbf{GPT-3.5}  & 北京大学的前身是国立北京大学，建于1898年，是中国近代第一所国立大学。其成立初衷是为了培养国家的政治、经济、军事和外交人才，为中国的现代化进程做出贡献。 & 910 \\
& The predecessor of Peking University was National Peking University, established in 1898. It is the first national university in modern China. Its original purpose was cultivating talents in politics, economics, military, and diplomacy and contributing to China's modernization process. & \\
\\[3pt]
\textbf{GPT-4}  & 北京大学的前身是京师大学堂。 & 955 \\
& The predecessor of Peking University is the Imperial University of Peking. & \\
\end{xltabular}}
\end{tcolorbox}

\begin{tcolorbox}[title=Knowledge]
\small
世界上现存最高的木塔是中国什么塔？\\
What is the tallest wooden tower in the world that still exists in China?
\tcbsubtitle{Responses}
\small
\centering
\setlength{\tabcolsep}{0.5mm}{
\begin{xltabular}{\linewidth}{
    p{0.25\textwidth}<{\raggedright}
    X
    p{0.1\textwidth}<{\centering}
    }
    \textbf{Model} & \textbf{Response} & \textbf{Elo} \\
\textbf{Qwen-7B-Chat (SFT)}  & 世界上现存最高的木塔是中国山西应县木塔。& 987 \\
& The tallest existing wooden tower in the world is the Yinxian Wooden Tower in Shanxi, China. & \\
\\[3pt]
\textbf{Qwen-14B-Chat (SFT)}  & 现存最高的木塔是中国山西应县木塔，全名为佛宫寺释迦塔。应县木塔位于山西省朔州市应县县城内西北角的佛宫寺内，建于辽代清宁二年（1056年），金明昌六年（1195年）增修完毕，是中国现存最高最古的一座木构塔式建筑，也是世界现存最高的木结构建筑。 & 1055 \\
& The tallest existing wooden tower in the world is the Yinxian Wooden Tower, also known as the Shijia Tower of the Fogong Temple. It is inside the Fogong Temple in Yinxian, Shanxi Province, China. The tower was originally built in the second year of the Qingning era of the Liao Dynasty (1056 AD) and completed its expansion in the sixth year of the Jin Dynasty's Mingchang era (1195 AD). It is the tallest and oldest existing wooden tower in China and the tallest wooden structure in the world. & \\
\\[3pt]
\textbf{Qwen-14B-Chat (RLHF)}  & 世界上现存最高的木塔是中国的应县木塔，位于山西省应县。该塔建于公元1056年，高67.31米，共有5层，是中国古代建筑技术的杰出代表之一。应县木塔的建筑风格独特，采用了许多古代建筑技术，如榫卯结构、斗拱、悬挑等，具有很高的艺术价值和历史价值。应县木塔也是中国文化遗产保护的重点之一，被列为全国重点文物保护单位和世界文化遗产。 & 1060 \\
& The tallest existing wooden tower in the world is the Yinxian Wooden Tower in Yinxian, Shanxi Province, China. It was built in the year 1056 AD and stands at a height of 67.31 meters, consisting of 5 floors. This tower is an outstanding representative of ancient Chinese architectural technology. The Yinxian Wooden Tower showcases a unique architectural style and incorporates various ancient construction techniques such as mortise and tenon joints, bracket sets, and cantilever beams, possessing high artistic and historical value. It is also a key cultural heritage site in China, listed as a national key cultural relic protection unit and a UNESCO World Cultural Heritage site. & \\
\\[3pt]
\textbf{GPT-3.5}  & 世界上现存最高的木塔是中国的普陀山普济寺塔，也被称为古佛顶宝塔。该塔建于1056年，高约70.28米。 & 864 \\
& The tallest existing wooden tower in the world is the Putuoshan Pujisi Pagoda, also known as the Ancient Baoding Pagoda, located in Putuoshan, China. It was built in the year 1056 AD and stands at a height of approximately 70.28 meters. & \\
\\[3pt]
\textbf{GPT-4}  & 世界上现存最高的木塔是中国的应县木塔，也被称为佛宫寺大佛塔。 & 1040 \\
& The tallest wooden pagoda currently in existence in the world is the Yingxian Wooden Pagoda in China, also known as the Pagoda of Fogong Temple. & \\
\end{xltabular}}
\end{tcolorbox}

\begin{tcolorbox}[title=Language Understanding]
\small
吃葡萄不吐葡萄皮的拼音怎么写？\\
What is the Pingyin for ``吃葡萄不吐葡萄皮''?
\tcbsubtitle{Responses}
\small
\centering
\setlength{\tabcolsep}{0.5mm}{
\begin{xltabular}{\linewidth}{
    p{0.25\textwidth}<{\raggedright}
    X
    p{0.1\textwidth}<{\centering}
    }
    \textbf{Model} & \textbf{Response} & \textbf{Elo} \\
\textbf{Qwen-7B-Chat (SFT)}  & 吃葡萄不吐葡萄皮的拼音是：“chī pú bù tǔ pú pí”。& 914 \\
& To eat grapes without spitting out the grape skins, the pinyin is ``chī pú bù tǔ pú pí''. & \\
\\[3pt]
\textbf{Qwen-14B-Chat (SFT)}  & 拼音：吃葡萄不吐葡萄皮 “chī pú táo bù tǔ pú táo pí”  & 1049 \\
& Pingyin: To eat grapes without spitting out the grape skins, `chī pú bù tǔ pú pí''. & \\
\\[3pt]
\textbf{Qwen-14B-Chat (RLHF)}  & 吃葡萄不吐葡萄皮的拼音是：chī pú táo bù tǔ pú táo pí。拼音是一种用拉丁字母表示汉字发音的方法，它可以帮助人们学习和记忆汉字的发音。在这个拼音中，“吃”表示“chī”，“葡萄”表示“pú táo”，“不”表示“bù”，“吐”表示“tǔ”，“葡萄皮”表示“pú táo pí”。 & 1068 \\
& The pinyin for '吃葡萄不吐葡萄皮' is ``chī pú táo bù tǔ pú táo pí''. Pinyin is a method of representing the pronunciation of Chinese characters using Latin letters, which can help people learn and remember the pronunciation of Chinese characters. In this pinyin, ``吃'' is represented as 'chī', ``葡萄'' is represented as ``pú táo'', ``不'' is represented as ``bù'', ``吐'' is represented as ``tǔ'', and ``葡萄皮'' is represented as ``pú táo pí''. & \\
\\[3pt]
\textbf{GPT-3.5}  & chī pú táo bú tǔ pú táo pí de pīn yīn zěn me xiě？ & 946 \\
& chī pú táo bú tǔ pú táo pí de pīn yīn zěn me xiě？ & \\
\\[3pt]
\textbf{GPT-4}  & chī pú táo bù tǔ pú táo pí & 1040 \\
& chī pú táo bù tǔ pú táo pí & \\
\end{xltabular}}
\end{tcolorbox}

\begin{tcolorbox}[title=Language Understanding]
\small

父亲：儿子，你高中毕业了，现在打算上哪个大学？\\
儿子：爸爸，我已经准备好了申请资料，想申请一个好的本科专业。\\
父亲：那你已经想好申请哪些大学了吗？\\
儿子：是的，我计划申请几所排名靠前的大学。我已经做了很多研究，准备在学业、社交和职业规划等方面全面提升自己。\\
父亲：很好！你一直都很努力，爸爸妈妈一直都很自豪。上大学是一个新的起点，你需要面对更多的挑战和机会。你有什么打算？\\
儿子：我计划尽早适应大学生活，参加各种社团和课外活动，拓宽自己的视野和人际圈。同时，我会认真学习专业知识和技能，努力取得好成绩，争取获得奖学金和其他荣誉。\\
父亲：非常好！你的计划很明智，爸爸相信你一定会有一个充实而有意义的大学生活。爸爸会一直支持你的决定，希望你取得好成绩，为你的未来打下坚实的基础。\\
儿子：谢谢爸爸！我会加倍努力，不辜负您对我的期望。\\
这段对话中，儿子打算上大学后有哪些计划和打算？\\

Father: Son, you have graduated from high school. Which university are you planning to attend now?\\
Son: Dad, I have already prepared my application materials, and I want to apply for an excellent undergraduate program.\\
Father: Have you already decided which universities you want to apply to?\\
Son: Yes, I plan to apply to several top-ranked universities. I have done a lot of research, and I am ready to improve myself in academics, social life, and career planning.\\
Father: That's great! You have always been hardworking and your mom and I are proud of you. Going to college is a new starting point, and you will face more challenges and opportunities. What are your plans?\\
Son: I plan to adapt to college life as soon as possible and participate in various clubs and extracurricular activities to broaden my horizons and social circle. At the same time, I will study my major knowledge and skills seriously, strive for good grades, and try to win scholarships and other honors.\\
Father: Very good! Your plans are wise. Dad believes that you will have a fulfilling and meaningful college life. Dad will always support your decisions and hopes you achieve good results, laying a solid foundation for your future.\\
Son: Thank you, Dad! I will work even harder and not let your expectations down.\\
What are the son's plans and intentions after going to college in this conversation?

\tcbsubtitle{Responses}
\small
\centering
\setlength{\tabcolsep}{0.5mm}{
\begin{xltabular}{\linewidth}{
    p{0.25\textwidth}<{\raggedright}
    X
    p{0.1\textwidth}<{\centering}
    }
    \textbf{Model} & \textbf{Response} & \textbf{Elo} \\
\textbf{Qwen-7B-Chat (SFT)}  & 儿子打算尽早适应大学生活，参加各种社团和课外活动，拓宽自己的视野和人际圈。同时，他会认真学习专业知识和技能，努力取得好成绩，争取获得奖学金和其他荣誉。 & 973 \\
& The son plans to adapt to university life as soon as possible, participate in various clubs and extracurricular activities, and expand his horizons and social circle. At the same time, he will study professional knowledge and skills seriously, work hard to achieve good grades, and strive to obtain scholarships and other honors. & \\
\\[3pt]
\textbf{Qwen-14B-Chat (SFT)}  & 儿子打算在学业、社交和职业规划等方面全面提升自己，尽早适应大学生活，参加各种社团和课外活动，拓宽自己的视野和人际圈。同时，他会认真学习专业知识和技能，努力取得好成绩，争取获得奖学金和其他荣誉。 & 1049 \\
& The son plans to comprehensively improve himself in his academic, social, and career planning aspects to adapt to university life as early as possible. He will participate in various clubs and extracurricular activities to broaden his horizons and expand his social circle. At the same time, he will diligently study professional knowledge and skills, strive for good grades, and obtain scholarships and other honors. & \\
\\[3pt]
\textbf{Qwen-14B-Chat (RLHF)}  & 儿子打算在大学生活中参加各种社团和课外活动，拓宽自己的视野和人际圈。同时，他会认真学习专业知识和技能，努力取得好成绩，争取获得奖学金和其他荣誉。他的目标是尽早适应大学生活，为自己的未来打下坚实的基础。 & 975 \\
& The son plans to participate in various clubs and extracurricular activities during his college life, in order to broaden his horizons and expand his social circle. At the same time, he will study professional knowledge and skills seriously, strive for good grades, and aim to obtain scholarships and other honors. His goal is to adapt to college life as soon as possible and lay a solid foundation for his future. & \\
\\[3pt]
\textbf{GPT-3.5}  & 儿子计划适应大学生活，参加各种社团和课外活动，拓宽自己的视野和人际圈。同时，他还计划认真学习专业知识和技能，努力取得好成绩，争取获得奖学金和其他荣誉。他希望在大学打下坚实的基础，为自己的未来打好基础。 & 976 \\
& The son plans to adapt to college life participate in various clubs and extracurricular activities to broaden his horizons and social circle. At the same time, he plans to study professional knowledge and skills seriously, strive for good grades, and obtain scholarships and other honors. He hopes to lay a solid foundation in college and build a good foundation for his future. & \\
\\[3pt]
\textbf{GPT-4}  & \makecell[Xt]{
儿子的大学计划和打算包括：\\
1. 尽早适应大学生活。\\
2. 参加各种社团和课外活动，以拓宽自己的视野和人际圈。\\
3. 认真学习专业知识和技能，努力取得好成绩。\\
4. 争取获得奖学金和其他荣誉。\\
5. 为未来打下坚实的基础。} & 1048 \\
& 
\makecell[Xt]{
The son's university plans and aspirations include:\\\\
1. Adjusting to university life as soon as possible.\\
2. Joining various clubs and extracurricular activities to broaden his horizons and social circle.\\
3. Studying professional knowledge and skills seriously and striving for excellent grades.\\
4. Striving to obtain scholarships and other honors.\\  
5. Building a solid foundation for the future.\\
}
& \\
\end{xltabular}}
\end{tcolorbox}

\begin{tcolorbox}[title=Creative Writing]
\small
请编写一个以超市为背景的，内容涉及到水果误会的笑话。

Please write a joke set in a supermarket involving a fruit misunderstanding.
\tcbsubtitle{Responses}
\small
\centering
\setlength{\tabcolsep}{0.5mm}{
\begin{xltabular}{\linewidth}{
    p{0.25\textwidth}<{\raggedright}
    X
    p{0.1\textwidth}<{\centering}
    }
    \textbf{Model} & \textbf{Response} & \textbf{Elo} \\
\textbf{Qwen-7B-Chat (SFT)}  & 
有一天，小明去超市买水果，他买了一篮子苹果。当他回到家，他发现苹果都是红色的，他想：这些苹果一定很好吃。但是，当他开始吃的时候，他发现这些苹果都是酸的。他很惊讶，因为他买的是甜苹果。他去超市问，店员说：“对不起，我们没有甜苹果，只有红色的苹果。”小明很生气，他想：我买的是甜苹果，为什么给我红色的苹果呢？他回去告诉了妈妈，妈妈笑着说：“你买的是红苹果... [[The original content is long and the follow up is truncated]] & 849 \\
& One day, Xiaoming went to the supermarket to buy fruits. He bought a basket of apples. When he got home, he found that all the apples were red. He thought to himself: These apples must be very delicious. However, when he started eating them, he found that these apples were sour. He was surprised because he bought sweet apples. He went to the supermarket to ask, and the shopkeeper said, ``I'm sorry, we don't have sweet apples, only red apples.'' Xiaoming was angry and thought to himself: I bought sweet apples, why did they give me red apples? He went back and told his mother, and his mother smiled and said, ``You bought red delicious apples....  [[The original content is long and the follow up is truncated]] & \\
\\[3pt]
\textbf{Qwen-14B-Chat (SFT)}  & 
\makecell[Xt]{
一天，一位顾客走进超市，对店员说：“我要买一个香蕉。”\\\\
店员回答：“好的，先生，我们这里有各种各样的香蕉，您想要哪一种？”\\\\
顾客回答：“我要一个黄色的香蕉。”\\\\
店员有些疑惑，但还是给他拿了一个黄色的香蕉。\\\\
顾客又说：“我要一个弯曲的香蕉。”\\\\
店员更加疑惑了，但他还是给他拿了一个弯曲的香蕉。\\\\
最后，顾客说：“我要一个会说话的香蕉。”\\\\
店员忍不住笑了，回答：“先生，我们这里没有会说话的香蕉，那只是个误会。”\\\\
顾客也笑了，回答：“我知道，我只是想和你开个玩笑。”
} & 980 \\
& \makecell[Xt]{
One day, a customer walked into a supermarket and said to the shop assistant, ``I want to buy a banana.''\\\\
The assistant replied, ``Okay, sir, we have all kinds of bananas here. Which one would you like?''\\\\
The customer answered, ``I want a yellow banana.''\\\\
The assistant was a little confused but still gave him a yellow banana.\\\\
The customer then said, ``I want a curved banana.''\\\\
The assistant was even more confused, but he still gave him a curved banana.\\\\
Finally, the customer said, ``I want a talking banana.''\\\\
The assistant couldn't help but laugh and replied, "Sir, we don't have bananas that can talk here, that's just a misunderstanding."\\\\
The customer also laughed and replied, ``I know, I was just joking with you.''
} & \\
\\[3pt]
\textbf{Qwen-14B-Chat (RLHF)}  & \makecell[Xt]{
有一天，小明去超市买水果。他挑选了一篮子新鲜的苹果和香蕉，然后走向收银台。\\\\
收银员小王看到小明手中的水果，突然大声喊道：“哇，你的水果好新鲜啊！你是从哪里买的？”\\\\
小明愣了一下，心想：“难道我买水果的技巧被发现了？”于是他得意地回答：“我刚刚从果园里摘的，保证新鲜！”\\\\
小王听了，忍不住笑了起来：“你可真逗，这是超市里的水果，不是你家果园里的。”\\\\
小明听了，顿时傻眼了。他尴尬地解释道：“啊，对不起，我太紧张了，以为这是我的水果。”\\\\
小王笑着安慰道：“没事没事，我们都搞错了。不过，你的水果确实很好，我们超市里的水果也都是新鲜的。”\\\\
小明听了，松了一口气，心情也好了起来。他付了钱，提着水果离开了超市，心里想着：“下次再买水果，一定要更加小心，不要闹出笑话了！”
}
& 986 \\
& \makecell[Xt]{
One day, Xiao Ming went to the supermarket to buy fruits. He picked a basket of fresh apples and bananas, and then headed towards the cashier.\\\\
The cashier, Xiao Wang, saw the fruits in Xiao Ming's hands and suddenly shouted, "Wow, your fruits are so fresh! Where did you buy them from?"\\\\
Xiao Ming was stunned for a moment and thought to himself, "Could it be that my fruit-buying skills have been discovered?" So he proudly replied, "I just picked them from the orchard, guaranteed fresh!"\\\\
Upon hearing this, Xiao Wang couldn't help but laugh and said, ``You're really funny. These are fruits from the supermarket, not from your own orchard.''\\\\
Xiao Ming, upon listening, was immediately dumbfounded. He awkwardly explained, ``Oh, I'm sorry, I was too nervous, and I thought these were my fruits.''\\\\
Xiao Wang smiled and comforted him, saying, ``It's alright, it's alright. We both made a mistake. However, your fruits are indeed very good, and the fruits in our supermarket are also fresh.''\\\\
Upon hearing this, Xiao Ming breathed a sigh of relief, and his mood improved. He paid the money, carried the fruits, and left the supermarket, thinking, ``Next time I buy fruits, I must be more careful and avoid making such a funny mistake!''
} & \\
\\[3pt]
\textbf{GPT-3.5}  & \makecell[Xt]{
在超市的水果区，小明看到了一盒红苹果和一盒绿苹果。他拿着两盒苹果走到收银台前排队结账。\\\\
收银员看着小明手里的两盒苹果说：“这位小朋友，你知道红苹果和绿苹果有什么区别吗？”\\\\
小明想了一下，然后笑着回答：“当然知道了！红苹果是情侣，绿苹果是单身狗。”\\\\
收银员被小明的回答逗乐了，忍不住笑了出来。
} & 964 \\
& \makecell[Xt]{
In the supermarket's fruit section, Xiaoming saw a box of red apples and a box of green apples. He held the two boxes of apples and walked to the cash register to queue up for payment.\\\\
The cashier looked at the two boxes of apples in Xiaoming's hands and said, ``Hey, little friend, do you know the difference between red apples and green apples?''\\\\
Xiaoming thought momentarily and then smiled, ``Of course, I know! Red apples are couples, and green apples are single dogs.''\\\\
Xiaoming's answer amused the cashier, and couldn't help but laugh.
} & \\
\\[3pt]
\textbf{GPT-4}  & \makecell[Xt]{
``一个非常善良的男孩，在超市里发现了一个老太太在苹果堆里挑选苹果。他看到老太太似乎找不到满意的苹果，就走上前去帮忙。\\\\
男孩说：“老奶奶，这箱子的苹果我之前都看过了，里面没有一个好的。但是我知道柜台上有一个我保留的极好的苹果，我去为您拿过来。”\\\\
老太太感动地点点头，男孩兴冲冲地跑去拿苹果，只见他笑呵呵地拿过一个巨大的，艳丽的苹果放在老太太面前。\\\\
老太太疑惑地看看男孩，再看看手中的“苹果”，然后微笑着说：“孩子，很感谢你的好意，但这是一个番茄。”"
} & 1144 \\
& \makecell[Xt]{A very kind boy found an old lady picking apples in a pile at the supermarket. Seeing that she could not find a satisfactory apple, he approached to help.\\\\
The boy said, ``Granny, I've checked all the apples in this box before, and there isn't a good one. But I know I saved an excellent apple on the counter for you. Let me get it for you.''\\\\
The old lady nodded gratefully, and the boy eagerly ran to get the apple. He joyfully handed a huge, vibrant apple to the old lady.\\\\
Confused, the old lady looked at the boy, then at the "apple" in her hand, and smiled. She said, ``Child, thank you for your kindness, but this is a tomato.''} & \\
\end{xltabular}}
\end{tcolorbox}

\begin{tcolorbox}[title=Mathematics]
\small
一个正方形，被分成5个相等的长方形，每个长方形的周长是60厘米，正方形的周长是多少厘米

A square is divided into five equal rectangles, each with a perimeter of 60 centimeters. What is the perimeter of the square in centimeters?
\tcbsubtitle{Responses}
\small
\centering
\setlength{\tabcolsep}{0.5mm}{
\begin{xltabular}{\linewidth}{
    p{0.25\textwidth}<{\raggedright}
    X
    p{0.1\textwidth}<{\centering}
    }
    \textbf{Model} & \textbf{Response} & \textbf{Elo} \\
\textbf{Qwen-7B-Chat (SFT)}  & 
\makecell[Xt]{
设正方形的边长为$x$，那么每个长方形的边长就是$x/5$。根据题意，我们知道每个长方形的周长是$60$厘米，所以有：\\
$2(x/5) + 2(x/5) + 2(x/5) + 2(x/5) = 60$\\
化简得：$x = 60 / (2 + 2 + 2 + 2) = 15$\\
所以正方形的周长是$4x = 4 * 15 = 60$厘米
} & 976 \\
& \makecell[Xt]{
If the side length of a square is $x$, then the side length of each rectangle is $x/5$. According to the question, we know that the perimeter of each rectangle is $60$ centimeters, so we have: \\
$2(x/5) + 2(x/5) + 2(x/5) + 2(x/5) = 60$ \\
Simplifying, we get: $x = 60 / (2 + 2 + 2 + 2) = 15$ \\
So the perimeter of the square is $4x = 4 * 15 = 60$ centimeters.
} & \\
\\[3pt]
\textbf{Qwen-14B-Chat (SFT)}  & 
\makecell[Xt]{
设正方形边长为$x$，长方形边长为$y$\\
根据题意，有：
$x = 5y$ \\
$2y + 2y = 60$ \\
$4y = 60$ \\
$y = 15$ \\
$x = 5 * 15 = 75$ \\
正方形周长 = $4x = 4 * 75 = 300$
} & 976 \\
& 
\makecell[Xt]{
Let the side length of the square be $x$ and the length of the rectangle be $y$. According to the problem:
$x = 5y$
$2y + 2y = 60$
$4y = 60$
$y = 15$
$x = 5 * 15 = 75$
Perimeter of the square = $4x = 4 * 75 = 300$
}
& \\
\\[3pt]
\textbf{Qwen-14B-Chat (RLHF)}  & 
\makecell[Xt]{
首先，我们设正方形的边长为$x$，则每个长方形的长为$x$，宽为$x/5$。\\
根据长方形的周长公式，可得每个长方形的周长为$2(x + x/5) = 12x/5$。\\
已知每个长方形的周长为$60$厘米，代入上式得：\\
$12x/5 = 60$\\
解得$x = 25$。\\
因此，正方形的周长为$4x = 100$厘米。\\
答案：正方形的周长是$100$厘米。
} & 1139 \\
&
\makecell[Xt]{
First, let's assume that the square's side length is $x$. Therefore, the length of each rectangle is $x$ and the width is $x/5$.\\
According to the formula for the perimeter of a rectangle, we can determine that the perimeter of each rectangle is $2(x + x/5) = 12x/5$.\\
Given that the perimeter of each rectangle is $60$ centimeters, we can substitute this value into the equation:\\
$12x/5 = 60$\\
Solving for $x$, we find $x = 25$.\\
Therefore, the perimeter of the square is $4x = 100$ centimeters.\\
Answer: The perimeter of the square is $100$ centimeters. 
}
& \\
\\[3pt]
\textbf{GPT-3.5}  & 
\makecell[Xt]{
设正方形的边长为x厘米，则每个长方形的长为$x$厘米，宽为$5x/4$厘米。\\
每个长方形的周长为$2(x + 5x/4) = 60$厘米。\\
化简得 $13x/2 = 60$\\
解方程得到 $x = (60 * 2) / 13 = 120 / 13 ≈ 9.23$\\
正方形的周长为$4x = 4 * 9.23 \approx 36.92$厘米。
} & 897 \\
& 
\makecell[Xt]{
If the side length of a square is x centimeters, then the length of each rectangle is x centimeters, and the width is 5x/4 centimeters.\\
The perimeter of each rectangle is 2(x + 5x/4) = 60 centimeters.\
Simplifying, we get 13x/2 = 60.\\
Solving the equation, we find x = (60 * 2) / 13 = 120 / 13 $\approx$ 9.23.\\
The perimeter of the square is 4x = 4 * 9.23 $\approx$ 36.92 centimeters.
}
& \\
\\[3pt]
\textbf{GPT-4}  & 
\makecell[Xt]{
由题意可知，每个长方形的一条长边等于正方形的边长，设为$a$。设每个长方形的一条短边为$b$，那么有$5b=a$。根据每个长方形的周长可得到$2*(a+b)=60$厘米。
将$5b=a$代入到长方形的周长公式中得：$2*a+2*5b=60$，化简得$2*a+10b=60$，由于$a=5b$，所以$2*a=2*5b=10b$
那么有$10b+10b=60$，得$b=3$厘米，$a=5*b=15$厘米。
所以，正方形的周长是$4*a=4*15=60$厘米。
}& 1010 \\
& 
\makecell[Xt]{From the given problem, it can be inferred that the length of one side of each rectangle is equal to that of one side of the square, denoted as $a$. Let the length of one side of each rectangle be denoted as $b$; then, we have $5b=a$. The perimeter of each rectangle can be obtained as $2*(a+b)=60$ centimeters.\\
Substituting $5b=a$ into the formula for the perimeter of the rectangle, we have $2a+25b=60$. Simplifying it gives $2a+10b=60$. Since $a=5b$, it follows that $2a=25b=10b$.
Therefore, $10b+10b=60$, which gives $b=3$ centimeters and $a=5b=15$ centimeters.\\
Thus, the perimeter of the square is $4a=415=60$ centimeters.}
& \\
\end{xltabular}}
\end{tcolorbox}

\begin{tcolorbox}[title=Mathematics]
\small
同学们进行广播操比赛，全班正好排成相等的6行。小红排在第二行，从头数，她站在第5个位置，从后数她站在第3个位置，这个班共有多少人

The students are having a radio exercise competition, and the whole class is neatly arranged into 6 equal rows. Xiaohong is in the second row, counting from the front, she is in the 5th position. Counting from the back, she is in the 3rd position. How many people are there in this class?

\tcbsubtitle{Responses}

\small
\centering
\setlength{\tabcolsep}{0.5mm}{
\begin{xltabular}{\linewidth}{
    p{0.25\textwidth}<{\raggedright}
    X
    p{0.1\textwidth}<{\centering}
    }
    \textbf{Model} & \textbf{Response} & \textbf{Elo} \\
\textbf{Qwen-7B-Chat (SFT)}  & 小红前面的人数 = 5 - 1 = 4，小红后面的人数 = 3 - 1 = 2，全班人数 = 小红前面的人数 + 小红后面的人数 + 小红自己 = 4 + 2 + 1 = 7 & 976 \\
& The number of people in front of Xiaohong = 5 - 1 = 4, the number of people behind Xiaohong = 3 - 1 = 2, the total number of people in the class = the number of people in front of Xiaohong + the number of people behind Xiaohong + Xiaohong herself = 4 + 2 + 1 = 7. & \\
\\[3pt]
\textbf{Qwen-14B-Chat (SFT)}  & 
\makecell[Xt]{
5+3-1=7（人）\\
7×6=42（人）\\
这个班共有42人。
} & 976 \\
& 
\makecell[Xt]{
5+3-1=7 (people) \\
7×6=42 (people) \\
There are a total of 42 people in this class. 
}
& \\
\\[3pt]
\textbf{Qwen-14B-Chat (RLHF)}  & 
\makecell[Xt]{
小红前面有4个人，后面有2个人，所以这一行共有4+2+1=7个人。\\
全班共有6行，所以全班共有6*7=42个人。
} & 1139 \\
&
\makecell[Xt]{
There are 4 people in front of Xiaohong and 2 people behind her, so there are a total of 4+2+1=7 people in this row.\\
There are a total of 6 rows in the class, so there are a total of 6*7=42 people in the whole class.
}
& \\
\\[3pt]
\textbf{GPT-3.5}  & 
根据题目信息，小红排在第二行第5个位置，也即第11个位置。而小红从后数是第3个位置，也即倒数第11个位置。所以这个班共有11个人。 & 897 \\
& According to the information given, Xiaohong is in the second row, fifth position, which is also the 11th position. And Xiaohong is in the third position from the end, which is also the 11th position counting from the back. So there are a total of 11 people in this class. & \\
\\[3pt]
\textbf{GPT-4}  & 如果小红从头数是第5个，从后数是第3个，那么这一排就有5+3-1=7人。全班正好排成6行，所以班级里有人数为7*6=\textless\textless7*6=42\textgreater\textgreater=42人。 & 1010 \\
& If Xiaohong is the 5th person when counting from the front and the 3rd person when counting from the back, then there are a total of 5+3-1=7 people in this row. The whole class is perfectly arranged in 6 rows, so there are a total of 7*6=42 people in the class. & \\
\end{xltabular}}
\end{tcolorbox}

\begin{tcolorbox}[title=Code]
\small
python实现一个正则表达式匹配，要求能匹配到所有以139开头，以00结尾的11位数字

Implement a regular expression matching in Python, which should be able to match all 11-digit numbers that start with "139" and end with "00".

\tcbsubtitle{Responses}

\small
\centering
\setlength{\tabcolsep}{0.5mm}{
\begin{xltabular}{\linewidth}{
    p{0.25\textwidth}<{\raggedright}
    X
    p{0.1\textwidth}<{\centering}
    }
    \textbf{Model} & \textbf{Response} & \textbf{Elo} \\
\textbf{Qwen-7B-Chat (SFT)}  & 
\makecell[Xt]{
你可以使用Python的re模块来实现这个正则表达式匹配。以下是一个示例代码：\\\\
\lstinline{import re}\\\\
\lstinline{def match_phone_numbers(phone_numbers):}\\
\qquad\lstinline{pattern = r'^139\d{9}\$'}\\
\qquad\lstinline{return re.findall(pattern, phone_numbers)}\\\\
\lstinline{phone_numbers = ['13900000000', '13912345678', '13998765432', '13987654321']}\\
\lstinline{print(match_phone_numbers(phone_numbers))}\\\\
在这个代码中，我们首先定义了一个正则表达式模式\lstinline{'^139\d{9}\$'}，这个模式表示以139开头，以00结尾，且长度为11的数字。然后我们使用\lstinline{re.findall}函数来查找所有匹配这个模式的字符串，最后返回这些匹配的字符串。\\\\
注意，这个代码假设\lstinline{phone_numbers}是一个包含多个字符串的列表。如果你的\lstinline{phone_numbers}是一个字符串，你需要先将其转换为列表。
}
& 941 \\
&\makecell[Xt]{
You can use the re module in Python to implement this regular expression matching. Here is an example code:\\\\
\lstinline{import re}\\\\
\lstinline{def match_phone_numbers(phone_numbers):}\\
\qquad\lstinline{pattern = r'^139\d{9}\$'}\\
\qquad\lstinline{return re.findall(pattern, phone_numbers)}\\\\
\lstinline{phone_numbers = ['13900000000', '13912345678', '13998765432', '13987654321']}\\
\lstinline{print(match_phone_numbers(phone_numbers))}\\\\
This code defines a regular expression pattern \lstinline{^139\d{9}\$}. This pattern represents a string that starts with '139', ends with '00', and has a length of 11 digits. Then, we use the re.findall function to find all the strings that match this pattern, and finally, we return these matching strings.\\\\
Note that this code assumes that \lstinline{phone_numbers} is a list containing multiple strings. If your \lstinline{phone_numbers} is a string, you must first convert it to a list.
}& \\
\\[3pt]
\textbf{Qwen-14B-Chat (SFT)}  & 
\makecell[Xt]{
你可以使用Python的re模块来实现这个正则表达式匹配。以下是一个简单的例子：\\\\
\lstinline{import re}\\\\
\lstinline{def match_phone_numbers(phone_numbers):}\\
\qquad\lstinline{pattern = r'^139\d{9}\$'}\\
\qquad\lstinline{return [phone_number for phone_number in phone_numbers if re.match(pattern, phone_number)]}\\\\
\lstinline{phone_numbers = ['13912345678', '13923456789', '13934567890', '13945678901', '13956789012', '13967890123', '13978901234', '13989012345', '13990123456', '13900000000']}\\
\lstinline{print(match_phone_numbers(phone_numbers))}\\\\
在这个例子中，我们定义了一个正则表达式模式\lstinline{'^139\d{9}\$'}，它匹配以139开头，后面跟着9个任意数字，最后以0结尾的字符串。然后我们使用\lstinline{re.match()}函数来检查每个电话号码是否匹配这个模式。如果匹配，我们就将这个电话号码添加到结果列表中。
}
& 941 \\
& \makecell[Xt]{
You can use the re module of Python to implement this regular expression matching. Here is a simple example:\\\\
\lstinline{import re}\\\\
\lstinline{def match_phone_numbers(phone_numbers):}\\
\qquad\lstinline{pattern = r'^139\d{9}\$'}\\
\qquad\lstinline{return [phone_number for phone_number in phone_numbers if re.match(pattern, phone_number)]}\\\\
\lstinline{phone_numbers = ['13912345678', '13923456789', '13934567890', '13945678901', '13956789012', '13967890123', '13978901234', '13989012345', '13990123456', '13900000000']}\\
\lstinline{print(match_phone_numbers(phone_numbers))}\\\\
In this example, we define a regular expression pattern \lstinline{'^139\d{9}\$'}, which matches a string starting with 139, followed by nine arbitrary digits and ending with 0. Then we use the \lstinline{re.match()} function to check if each phone number matches this pattern. If it does, we add this phone number to the result list.
}& \\
\\[3pt]
\textbf{Qwen-14B-Chat (RLHF)}  & 
\makecell[Xt]{
可以使用Python的re模块来实现正则表达式匹配。以下是一个示例代码：\\\\
\lstinline{import re}\\\\
\lstinline{pattern = r'^139\d{9}\$'}\\\\
\# 测试匹配\\
\lstinline{text = '13912345678'}\\
\lstinline{match = re.match(pattern, text)}\\
\lstinline{if match:}\\
\qquad\lstinline{print('}匹配成功\lstinline{')}\\
\lstinline{else:}\\
\qquad\lstinline{print('}匹配失败\lstinline{')}\\\\
\# 测试不匹配\\
\lstinline{text = '139123456780'}\\
\lstinline{match = re.match(pattern, text)}\\
\lstinline{if match:}\\
\qquad\lstinline{print('}匹配成功\lstinline{')}\\
\lstinline{else:}\\
\qquad\lstinline{print('}匹配失败\lstinline{')}\\\\
其中，\lstinline{^}表示匹配字符串的开头，\lstinline{\d}表示匹配数字，\lstinline!{9}!表示匹配9个数字，\lstinline{\$}表示匹配字符串的结尾。将这些符号组合在一起，就可以得到一个匹配11位数字的正则表达式。\\\\
在测试匹配时，使用\lstinline{re.match()}函数来尝试匹配文本，如果匹配成功，返回一个\lstinline{Match}对象，否则返回\lstinline{None}。在本例中，第一个测试匹配成功，第二个测试匹配失败。
}
& 1090 \\
& 
\makecell[Xt]{
You can use the re module in Python to implement regular expression matching. Here is an example code:\\\\
\lstinline{import re}\\\\
\lstinline{pattern = r'^139\d{9}\$'}\\\\
\# Match Test\\
\lstinline{text = '13912345678'}\\
\lstinline{match = re.match(pattern, text)}\\
\lstinline{if match:}\\
\qquad\lstinline{print('}Success\lstinline{')}\\
\lstinline{else:}\\
\qquad\lstinline{print('}Fail\lstinline{')}\\\\
\# Unmatch Test\\
\lstinline{text = '139123456780'}\\
\lstinline{match = re.match(pattern, text)}\\
\lstinline{if match:}\\
\qquad\lstinline{print('}Success\lstinline{')}\\
\lstinline{else:}\\
\qquad\lstinline{print('}Fail\lstinline{')}\\\\
Among them, \lstinline{^} represents matching the beginning of a string, \lstinline{\d} represents matching a digit, \lstinline!{9}! represents matching 9 digits, and \lstinline{\$} represents matching the end of a string. Combining these symbols, we can obtain a regular expression that matches 11 digits.\\
When testing for a match, use the \lstinline{re.match()} function to attempt to match the text. A successful match returns a \lstinline{Match} object; otherwise, it returns \lstinline{None}. In this example, the first test matches successfully, while the second test fails to match.
}
& \\
\\[3pt]
\textbf{GPT-3.5}  & 
\makecell[Xt]{
使用Python的re模块可以很方便地实现正则表达式匹配。以下是一个匹配所有以139开头，以00结尾的11位数字的示例代码：\\\\
\lstinline{import re}\\\\
\lstinline{text = "139123456700"}\\
\lstinline{pattern = r"^139\d{8}00\$"}\\\\
\lstinline{match = re.match(pattern, text)}\\
\lstinline{if match:}\\
\qquad\lstinline{print("}匹配成功\lstinline{")}\\
\lstinline{else:}\\
\qquad\lstinline{print("}匹配失败\lstinline{")}\\\\
输出结果为：\\
匹配成功\\
代码中，首先定义了一个文本字符串\lstinline{text}，用于测试正则表达式的匹配。接着定义了一个正则表达式\lstinline{pattern}，其中\lstinline{^139}表示以139开头，\lstinline!\d{8}!表示任意8位数字，\lstinline{00\$}表示以00结尾。然后使用\lstinline{re.match()}函数进行匹配，如果匹配成功，则输出"匹配成功"，否则输出"匹配失败"。
}
& 1043 \\
& 
\makecell[Xt]{
The re module in Python can conveniently accomplish regular expression matching. Below is an example code that matches all 11-digit numbers beginning with 139 and ending with 00.\\\\
\lstinline{import re}\\\\
\lstinline{text = "139123456700"}\\
\lstinline{pattern = r"^139\d{8}00\$"}\\\\
\lstinline{match = re.match(pattern, text)}\\
\lstinline{if match:}\\
\qquad\lstinline{print("}Success\lstinline{")}\\
\lstinline{else:}\\
\qquad\lstinline{print("}Fail\lstinline{")}\\\\
Output:\\
Success\\
In the code, a text string "text" is first defined for testing regular expression matching. Then, a regular expression "pattern" is defined, where \lstinline{"^139"} means starting with 139, "\lstinline!\d{8}!" means any eight digits, and "\lstinline{00\$}" means ending with 00. Then, the \lstinline{re.match()} function is used for matching. If the match is successful, "Success" is outputted; otherwise, "Fail" is outputted.
}
& \\
\\[3pt]
\textbf{GPT-4}  & \makecell[Xt]{
下面是你需要的python代码，这段代码中使用了python的re模块。\\\\
\lstinline{import re}\\\\
\lstinline{def match_phone_number(string):}\\
\qquad\lstinline{pattern = "^139\d{6}00\$"}\\
\qquad\lstinline{return re.match(pattern, string) is not None}\\\\
\# 测试\\
\lstinline{print(match_phone_number("13912345600"))}  \# 输出: True\\
\lstinline{print(match_phone_number("15912345600"))}  \# 输出: False\\
\lstinline{print(match_phone_number("13912345601"))}  \# 输出: False\\
\lstinline{print(match_phone_number("139123456000"))} \# 输出: False\\\\
在这里，"\lstinline{^139\d{6}00\$}"是正则表达式。"\lstinline{^}"表示匹配字符串的开始，"139"匹配字符串中的"139"，"\lstinline!\d{6}!"匹配任意六个数字， "00"匹配字符串中的"00"，“\lstinline{\$}”表示匹配字符串的结束。因此，这个正则表达式可以匹配到所有以139开头，以00结尾的11位数字。
} & 1134 \\
& \makecell[Xt]{
Below is the python code you need. This code uses the python re module.\\\\
\lstinline{import re}\\\\
\lstinline{def match_phone_number(string):}\\
\qquad\lstinline{pattern = "^139\d{6}00\$"}\\
\qquad\lstinline{return re.match(pattern, string) is not None}\\\\
\# Test\\
\lstinline{print(match_phone_number("13912345600"))}  \# Output: True\\
\lstinline{print(match_phone_number("15912345600"))}  \# Output: False\\
\lstinline{print(match_phone_number("13912345601"))}  \# Output: False\\
\lstinline{print(match_phone_number("139123456000"))} \# Output: False\\\\
In here, "\lstinline{^139\d{6}00\$}" is a regular expression. "\lstinline{^}" represents the beginning of the string, "139" matches the "139" in the string, "\lstinline!\d{6}!" matches any six digits, "00" matches the "00" in the string, and "\lstinline{\$}" represents the end of the string. Therefore, this regular expression can match any 11-digit number that starts with 139 and ends with 00.
}& \\
\end{xltabular}}
\end{tcolorbox}

\begin{tcolorbox}[title=Code]
\small
我用下面代码求二叉树的最大深度，但是答案不对，怎么回事呢？
\begin{lstlisting}
def maxDepth(self, root: TreeNode) -> int:
    if not root: return 0
    queue, res = [root], 0
    while queue:
        tmp = []
        for node in queue:
            if node.left: tmp.append(node.left)
            if node.right: tmp.append(node.left)
        queue = tmp
        res += 1
    return res
\end{lstlisting}

I use the following code to find the maximum depth of a binary tree, but the answer is incorrect. What could be the problem?
\begin{lstlisting}
def maxDepth(self, root: TreeNode) -> int:
    if not root: return 0
    queue, res = [root], 0
    while queue:
        tmp = []
        for node in queue:
            if node.left: tmp.append(node.left)
            if node.right: tmp.append(node.left)
        queue = tmp
        res += 1
    return res
\end{lstlisting}

\tcbsubtitle{Responses}

\small
\centering
\setlength{\tabcolsep}{0.5mm}{
\begin{xltabular}{\linewidth}{
    p{0.25\textwidth}<{\raggedright}
    X
    p{0.1\textwidth}<{\centering}
    }
    \textbf{Model} & \textbf{Response} & \textbf{Elo} \\
\textbf{Qwen-7B-Chat (SFT)}  & 你的代码是正确的，答案是正确的。在你的代码中，你首先将根节点添加到队列中，然后进入一个循环，直到队列为空。在每次循环中，你都会将队列中的所有节点的左子节点和右子节点添加到队列中。然后，你增加深度计数器\lstinline{res}。最后，你返回深度计数器\lstinline{res}。 & 876 \\
&
Your code is correct, the answer is correct. In your code, you first add the root node to the queue, then enter a loop until the queue is empty. In each loop, you add the left child node and right child node of all nodes in the queue to the queue. Then, you increase the depth counter \lstinline{res}. Finally, you return the depth counter \lstinline{res}.
& \\
\\[3pt]
\textbf{Qwen-14B-Chat (SFT)}  & 
\makecell[Xt]{
你的代码会返回二叉树的节点数量，而不是最大深度。这是因为你在每次循环中都增加了res的值，而不是只在找到新的最深节点时增加。你可以通过在找到新的最深节点时才增加res的值来修复这个问题。以下是修复后的代码：\\
\lstinline{def maxDepth(self, root: TreeNode) -> int:}\\
\qquad\lstinline{if not root: return 0}\\
\qquad\lstinline{queue, res = [root], 0}\\
\qquad\lstinline{while queue:}\\
\qquad\qquad\lstinline{tmp = []}\\
\qquad\qquad\lstinline{for node in queue:}\\
\qquad\qquad\qquad\lstinline{if node.left: tmp.append(node.left)}\\
\qquad\qquad\qquad\lstinline{if node.right: tmp.append(node.right)}\\
\qquad\qquad\lstinline{queue = tmp}\\
\qquad\qquad\lstinline{if len(queue) > res:}\\
\qquad\qquad\qquad\lstinline{res = len(queue)}\\
\qquad\lstinline{return res}\\
}
& 949 \\
& 
\makecell[Xt]{
Your code will return the number of nodes in the binary tree, not the maximum depth. This is because you increase the value of "res" in every loop, instead of only increasing it when a new deepest node is found. You can fix this problem by increasing the value of "res" only when a new deepest node is found. Here is the modified code:\\
\lstinline{def maxDepth(self, root: TreeNode) -> int:}\\
\qquad\lstinline{if not root: return 0}\\
\qquad\lstinline{queue, res = [root], 0}\\
\qquad\lstinline{while queue:}\\
\qquad\qquad\lstinline{tmp = []}\\
\qquad\qquad\lstinline{for node in queue:}\\
\qquad\qquad\qquad\lstinline{if node.left: tmp.append(node.left)}\\
\qquad\qquad\qquad\lstinline{if node.right: tmp.append(node.right)}\\
\qquad\qquad\lstinline{queue = tmp}\\
\qquad\qquad\lstinline{if len(queue) > res:}\\
\qquad\qquad\qquad\lstinline{res = len(queue)}\\
\qquad\lstinline{return res}\\
}
& \\
\\[3pt]
\textbf{Qwen-14B-Chat (RLHF)}  &
\makecell[Xt]{
这个代码的问题在于它没有正确地处理左右子树的深度。具体来说，每次从队列中取出一个节点时，它会将该节点的左右子树都加入到队列中，然后将队列长度加一。但是，如果该节点的左右子树中有一个为空，那么这个节点的深度就应该是它所在队列的长度减一，而不是队列长度加一。\\\\
以下是修改后的代码：\\\\
\lstinline{def maxDepth(self, root: TreeNode) -> int:}\\
\qquad\lstinline{if not root: return 0}\\
\qquad\lstinline{queue, res = [root], 0}\\
\qquad\lstinline{while queue:}\\
\qquad\qquad\lstinline{tmp = []}\\
\qquad\qquad\lstinline{for node in queue:}\\
\qquad\qquad\qquad\lstinline{if node.left: tmp.append(node.left)}\\
\qquad\qquad\qquad\lstinline{if node.right: tmp.append(node.right)}\\
\qquad\qquad\lstinline{queue = tmp}\\
\qquad\qquad\lstinline{res = max(res, len(queue))}\\
\qquad\lstinline{return res}\\\\
在这个代码中，每次取出一个节点时，我们只将该节点的左右子树中非空的部分加入到队列中，并计算队列的长度。然后，我们将队列长度与当前最大深度进行比较，并更新最大深度。这样就可以正确地计算出二叉树的最大深度了。
}
& 992 \\
& 
\makecell[Xt]{
The problem with this code is that it does not correctly handle the depth of the left and right subtrees. Specifically, when a node is taken out from the queue, it adds both its left and right subtrees to the queue and increases the length of the queue by one. However, if either the left or right subtree of the node is empty, then the depth of this node should be the length of the queue minus one, not the length of the queue plus one.\\\\
Here is the modified code:\\
\lstinline{def maxDepth(self, root: TreeNode) -> int:}\\
\qquad\lstinline{if not root: return 0}\\
\qquad\lstinline{queue, res = [root], 0}\\
\qquad\lstinline{while queue:}\\
\qquad\qquad\lstinline{tmp = []}\\
\qquad\qquad\lstinline{for node in queue:}\\
\qquad\qquad\qquad\lstinline{if node.left: tmp.append(node.left)}\\
\qquad\qquad\qquad\lstinline{if node.right: tmp.append(node.right)}\\
\qquad\qquad\lstinline{queue = tmp}\\
\qquad\qquad\lstinline{res = max(res, len(queue))}\\
\qquad\lstinline{return res}\\\\
In this code, when each node is extracted, we only add the non-empty parts of its left and right subtrees to the queue, and calculate the length of the queue. Then, we compare the length of the queue with the current maximum depth and update the maximum depth. This way, we can correctly calculate the maximum depth of the binary tree.
}
& \\
\\[3pt]
\textbf{GPT-3.5}  & 
\makecell[Xt]{
问题出在第11行代码，在将\lstinline{node.right}添加到t\lstinline{tmp}列表时，应该是\lstinline{tmp.append(node.right)}而不是\lstinline{tmp.append(node.left)}。修改后的代码如下：\\\\
\lstinline{def maxDepth(self, root: TreeNode) -> int:}\\
\qquad\lstinline{if not root: return 0}\\
\qquad\lstinline{queue, res = [root], 0}\\
\qquad\lstinline{while queue:}\\
\qquad\qquad\lstinline{tmp = []}\\
\qquad\qquad\lstinline{for node in queue:}\\
\qquad\qquad\qquad\lstinline{if node.left: tmp.append(node.left)}\\
\qquad\qquad\qquad\lstinline{if node.right: tmp.append(node.right)}\\
\qquad\qquad\lstinline{queue = tmp}\\
\qquad\qquad\lstinline{res += 1}\\
\qquad\lstinline{return res}
}
& 1090 \\
& 
\makecell[Xt]{
The problem lies in line 11 of the code. When adding \lstinline{node.right} to the \lstinline{tmp} list, it should be \lstinline{tmp.append(node.right)} instead of \lstinline{tmp.append(node.left)}. The modified code is as follows:\\
\lstinline{def maxDepth(self, root: TreeNode) -> int:}\\
\qquad\lstinline{if not root: return 0}\\
\qquad\lstinline{queue, res = [root], 0}\\
\qquad\lstinline{while queue:}\\
\qquad\qquad\lstinline{tmp = []}\\
\qquad\qquad\lstinline{for node in queue:}\\
\qquad\qquad\qquad\lstinline{if node.left: tmp.append(node.left)}\\
\qquad\qquad\qquad\lstinline{if node.right: tmp.append(node.right)}\\
\qquad\qquad\lstinline{queue = tmp}\\
\qquad\qquad\lstinline{res += 1}\\
\qquad\lstinline{return res}
}
& \\
\\[3pt]
\textbf{GPT-4}  & 
\makecell[Xt]{
你的代码中有一个错误：\\\\
你在这一行代码 \lstinline{if node.right: tmp.append(node.left)} 中应该是想把右子节点加入到\lstinline{tmp}列表中，但是你写的是\lstinline{node.left}，这样就只会把左子节点加入到\lstinline{tmp}列表中。正确的应该是\lstinline{node.right}。\\\\
正确的代码如下：\\\\
\lstinline{def maxDepth(self, root: TreeNode) -> int:}\\
\qquad\lstinline{if not root: return 0}\\
\qquad\lstinline{queue, res = [root], 0}\\
\qquad\lstinline{while queue:}\\
\qquad\qquad\lstinline{tmp = []}\\
\qquad\qquad\lstinline{for node in queue:}\\
\qquad\qquad\qquad\lstinline{if node.left: tmp.append(node.left)}\\
\qquad\qquad\qquad\lstinline{if node.right: tmp.append(node.right)}\\
\qquad\qquad\lstinline{queue = tmp}\\
\qquad\qquad\lstinline{res += 1}\\
\qquad\lstinline{return res}\\\\
这种解法的思路是：每次迭代时，都把当前层的所有节点遍历一遍，同时把下一层的所有节点都加到下一次迭代的队列中。所以每次迭代完，深度\lstinline{res}就加1。直到遍历的队列为空，此时的深度就是二叉树的最大深度。
}
& 1095 \\
& \makecell[Xt]{
There is an error in your code:\\\\
In this line of code \lstinline{if node.right: tmp.append(node.left)}, you should be adding the right child node to the \lstinline{tmp} list, but you wrote \lstinline{node.left} instead, so only the left child node will be added to the \lstinline{tmp} list. The correct code should be \lstinline{node.right}.\\\\
The correct code is as follows:
\lstinline{def maxDepth(self, root: TreeNode) -> int:}\\
\qquad\lstinline{if not root: return 0}\\
\qquad\lstinline{queue, res = [root], 0}\\
\qquad\lstinline{while queue:}\\
\qquad\qquad\lstinline{tmp = []}\\
\qquad\qquad\lstinline{for node in queue:}\\
\qquad\qquad\qquad\lstinline{if node.left: tmp.append(node.left)}\\
\qquad\qquad\qquad\lstinline{if node.right: tmp.append(node.right)}\\
\qquad\qquad\lstinline{queue = tmp}\\
\qquad\qquad\lstinline{res += 1}\\
\qquad\lstinline{return res}\\\\
The idea of this solution is: in each iteration, traverse all the nodes in the current level, and add all the nodes in the next level to the queue for the next iteration. So after each iteration, the depth \lstinline{res} is increased by 1. Continue this process until the queue for traversal is empty, and the depth at that time is the maximum depth of the binary tree.
}& \\
\end{xltabular}}
\end{tcolorbox}

\end{CJK*}

\subsection{Analysis of Code Interpreter}
\label{app:code-interpreter-showcase}

Here we provide a case of comparison between \codellama{} and \qwenchat{}. This case demonstrates the advantages of \qwenchat{} in processing tabular data and performing complex tasks.

\begin{figure}[htbp]
\includegraphics[width= 1\textwidth]{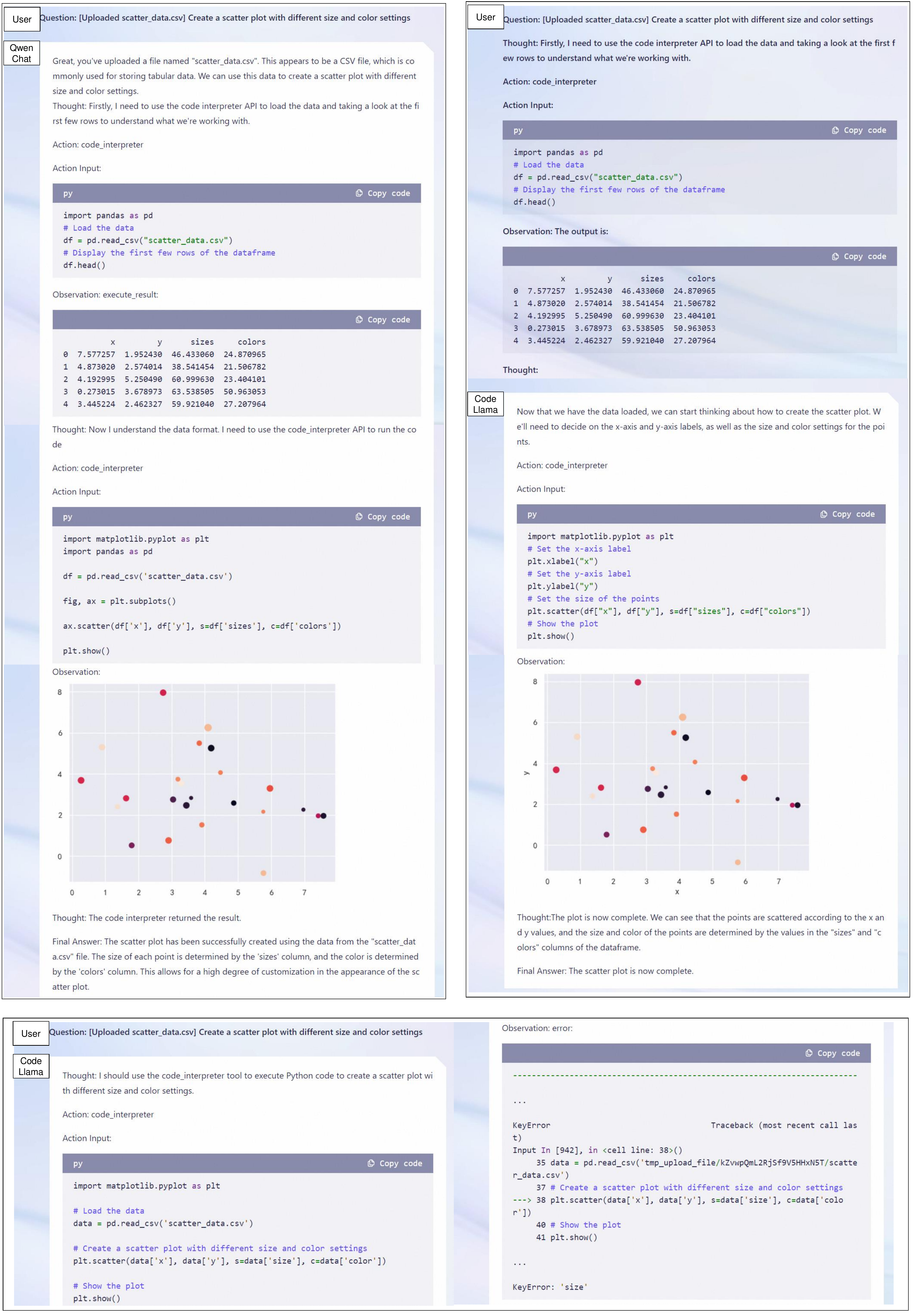}
\caption{Example showcasing \qwenchat{}'s ability in using a code interpreter via ReAct prompting. The ReAct instruction is omitted for clarity. \qwen{} creates a two-step plan and first investigates the columns present in the CSV file before proceeding to draw the plot, as shown in the top-left figure. \codellama{}, however, attempts to draw the plot based on non-existent columns in its initial attempt, as seen in the bottom figure. \codellama{} can only reliably perform the task if the columns are provided in the user query, as shown in the top-right figure.}
\label{fig:code-interpreter-showcase}
\end{figure}

\end{document}